\documentclass{article}

\usepackage{arxiv}

\usepackage[utf8]{inputenc} 
\usepackage[T1]{fontenc}    
\usepackage{hyperref}       
\usepackage{url}            
\usepackage{booktabs}       
\usepackage{amsfonts}       
\usepackage{nicefrac}       
\usepackage{microtype}      
\usepackage{lipsum}
\usepackage{multirow}
\usepackage{caption}
\usepackage{float}
\usepackage{fancyhdr}
\usepackage{tikz}
\usetikzlibrary{arrows.meta,positioning,fit,shapes.multipart,shapes.misc,calc}
\usepackage{geometry}
\usepackage{pgffor}
\usepackage{esvect}
\usepackage{graphicx}
\usepackage{amsmath}
\usepackage{hyperref}
\usepackage{xcolor}   
\usepackage{tabularx}
\usepackage[frozencache,cachedir=.]{minted}
\graphicspath{ {./images/} }

\title{FABRIC: \textbf{F}ramework for \textbf{A}gent-\textbf{B}ased \textbf{R}ealistic \textbf{I}ntelligence \textbf{C}reation \\
\large Weaving Synthetic Enterprise Data for Training Autonomous Agents
}

\author{
 Abhigya Verma\thanks{Equal contribution.} \\
 ServiceNow \\
 \texttt{abhigya.verma@servicenow.com} \\
   \And
 Seganrasan Subramanian\footnotemark[1] \\
 ServiceNow \\
 \texttt{seganrasan.subramanian@servicenow.com} \\
 \And
 Nandhakumar Kandasamy\footnotemark[1] \\
 ServiceNow \\
 \texttt{nandhakumar.k@servicenow.com} \\
  \And
 Naman Gupta\footnotemark[1] \\
 ServiceNow \\
 \texttt{naman.gupta@servicenow.com} \\
}

\begin{document}

\maketitle

\begin{abstract}
Large language models (LLMs) are increasingly deployed as \emph{agents}, expected to decompose goals, invoke tools, and verify results in dynamic environments. Realizing these capabilities requires access to \emph{agentic data}\cite{plaat2025agentic}—structured interaction records that couple user intents with tool specifications, argument-grounded calls, and verifiable execution traces. However, collecting such data from human annotators is costly, time-consuming, and difficult to scale.

We present a unified framework for synthesizing agentic data using only LLMs, without any human-in-the-loop supervision. This framework decomposes generation into modular pipelines that produce complete interaction records spanning task specifications, tool definitions, policy pseudocode, natural language exchanges, and execution traces. Records conform to strict syntactic and semantic constraints, ensuring machine-parseability and faithful alignment across inputs, outputs, and tool calls.

Beyond single tasks, there is support for both multi-task and multi-turn agent interactions, enabling the construction of datasets that reflect the full spectrum of tool-use competencies. To ensure quality and consistency, the framework integrates constrained generation formats, JSON-schema validation, and judge-based filtering.

This paper formalizes the schema for agentic records, details the prompt design principles that guide generation, and introduces scalable pipelines for high-quality synthetic data. By providing a reproducible, LLM-only alternative to manual collection, hence advancing the development of agentic LLMs capable of robust tool use.
\end{abstract}



\section{Introduction}
\label{sec:intro}

Large language models (LLMs) are increasingly deployed as \emph{agents} that decompose goals, invoke external tools, and coordinate multi-step workflows \cite{andreas2022language}. Realizing these capabilities requires access to \emph{agentic data}---structured, machine-interpretable records that capture the full trajectory of tool-augmented task execution, including user intent, tool specifications, decision logic, and verifiable execution traces.

\subsection{Motivation}
Agentic data differs fundamentally from conventional text corpora. Whereas standard datasets emphasize linguistic fidelity, agentic datasets must encode executable actions. Each record specifies not only natural language exchanges but also function calls, argument bindings, and execution outcomes. Such structured supervision is essential for enabling LLMs to learn robust tool-usage policies and for evaluating them on correctness, alignment, and reproducibility.

\subsection{Challenges in Data Acquisition}
Manual collection of agentic data at scale is infeasible. Human annotators must understand APIs, generate type-correct arguments, and validate tool responses across domains, a process that is costly, error-prone, and difficult to scale. Surveys of synthetic data generation \cite{chang2024survey, bauer2024comprehensive, jordon2024synthetic} highlight the benefits of automatic pipelines: scalability, diversity, and cost reduction. However, agentic settings impose stricter requirements due to the complexity of reasoning, planning, and execution fidelity \cite{sapkota2025agentic, wu2025agentic}.

\subsection{Requirements for Synthetic Agentic Data}
From prior work and system analyses, we identify the following requirements for automatically generated agentic data:
\begin{itemize}
  \item \textbf{Logical consistency}: argument values must respect type and semantic constraints, and call sequences must preserve dependency structure.
  \item \textbf{Execution fidelity}: traces must be verifiable and correspond to replayable executions.
  \item \textbf{Schema validity}: outputs must conform to machine-parseable schemas with strict I/O alignment.
  \item \textbf{Diversity and coverage}: datasets should include varied workflows, counterfactuals, and edge cases to improve robustness.
  \item \textbf{Scalability through automation}: reliance on human-in-the-loop supervision must be eliminated to achieve sufficient scale.
\end{itemize}

\subsection{Research Questions}
This work is guided by three research questions:
\begin{enumerate}
    \item How can an LLM-only framework generate synthetic agentic data that meets structural and semantic requirements without human supervision?
    \item Which modular pipelines best span the granularity of agentic supervision, from atomic function calls to long-horizon, multi-turn tasks?
    \item How can automated validation and schema enforcement guarantee that generated records are directly usable for training and evaluation?
\end{enumerate}

We propose a unified framework \emph{SYTHIA: \textbf{S}ynthetic \textbf{Y}ielding of \textbf{T}raining data for \textbf{H}uman-\textbf{I}ndependent \textbf{A}gents}. The framework consists of four modular pipelines, each designed to generate a distinct class of agentic data.

\section{Related Work}
\label{sec:related}

Recent advances in large language model (LLM) agents have spurred research across four interrelated areas: (1) simulation environments for realistic agentic behavior, (2) benchmarks for tool-augmented reasoning and evaluation, (3) methods for constructing synthetic agentic datasets, and (4) frameworks for tool-use, function-calling, and structured execution traces. We review representative efforts in each category and situate our framework relative to these lines of work.

\begin{table}[!htbp]
\centering
\begin{tabular}{p{3.5cm}p{1cm}p{11cm}}
\toprule
\textbf{Name} & \textbf{Year} & \textbf{Description} \\
\midrule
\textit{Self-Instruct}\cite{wang2022selfinstruct} & 2022 & LLM-only instruction synthesis; seed for fully synthetic data pipelines. \\ \hline
\textit{Mind2Web}\cite{deng2023mind2web} & 2023 & Human-annotated user trajectories over real websites for generalist agent training. \\ \hline
\textit{AgentBench}\cite{liu2023agentbench} & 2023 & Multi-domain agentic benchmark across reasoning and interaction settings. \\ \hline
\textit{AutoGen}\cite{wu2023autogen} & 2023 & Multi-agent conversational framework for synthetic data generation. \\ \hline
\textit{CAMEL}\cite{li2023camel} & 2023 & Communicative self-play agents for role-driven synthetic data. \\ \hline
\textit{ToolCoder}\cite{zhang2023toolcoder} & 2023 & Teaches code generation models to use API tools with schema alignment. \\ \hline
\textit{TaoBench}\cite{yang2023taobench} & 2023 & Task-oriented benchmark for evaluating LLM agents on multi-step tool-use and reasoning tasks. \\ \hline
\textit{Magpie}\cite{xu2024magpie} & 2024 & Alignment data generation from scratch with minimal inputs. \\ \hline
\textit{UltraFeedback}\cite{cui2024ultrafeedback} & 2024 & Large-scale preference data via AI feedback loops. \\ \hline
\textit{Arena Learning}\cite{luo2024arena} & 2024 & Iterative post-training data flywheel based on simulated chatbot arenas. \\ \hline
\textit{1B Personas}\cite{ge2024scaling} & 2024 & Large-scale persona-driven synthetic data generation. \\ \hline
\textit{LLM2LLM}\cite{lee2024llm2llm} & 2024 & Iterative refinement of data via LLM-to-LLM enhancement. \\ \hline
\textit{WebArena}\cite{zhou2024webarena} & 2024 & Realistic web environment with functional sites and long-horizon agent tasks. \\ \hline
\textit{BrowserGym}\cite{chezelles2024browsergym} & 2024 & Gym-style framework and benchmark suite for web-based agent training. \\ \hline
\textit{AppWorld}\cite{trivedi2024appworld} & 2024 & Multi-app environment for interactive code-based agent workflows. \\ \hline
\textit{WorkArena}\cite{drouin2024workarena} & 2024 & Evaluation framework on production websites for task-solving agents. \\ \hline
\textit{ST-WebAgentBenchh}\cite{levy2024stwebagentbench} & 2024 & Evaluates web agent safety and trustworthiness in open-ended settings. \\ \hline
\textit{AgentInstruct}\cite{mitra2024agentinstruct} & 2024 & Agentic flows for generative teaching and data creation. \\ \hline
\textit{AgentTrek}\cite{xu2025agenttrek} & 2025 & Converts web tutorials into executable agent trajectories. \\ \hline
\textit{BFCL}\cite{patil2025bfcl} & 2025 & Leaderboard for structured function-calling accuracy and argument correctness. \\
\bottomrule
\end{tabular}
\caption{Representative works across environments, benchmarks, synthetic data generation, and structured tool-use.}
\label{tab:agentic}
\end{table}

\subsection{Web-Based and Interactive Agent Environments}

Simulation environments provide reproducible testbeds for evaluating autonomous agents.  
\textbf{WebArena}\cite{zhou2024webarena}, \textbf{BrowserGym}\cite{chezelles2024browsergym}, and \textbf{WorkArena}\cite{drouin2024workarena} target realistic web tasks, offering functional sites and long-horizon goals.  
\textbf{AppWorld}\cite{trivedi2024appworld} generalizes this paradigm to multi-app ecosystems with user agents, supporting code-rich workflows.  
These environments support controlled evaluation but remain limited by dependence on pre-built platforms and runtime execution.

\subsection{Benchmarks for Tool-Augmented Agents}

Benchmarks assess the reasoning, tool-use, and safety of language agents.  
\textbf{AgentBench}\cite{liu2023agentbench} measures multi-domain decision-making abilities across interactive tasks.  
The \textbf{Berkeley Function-Calling Leaderboard (BFCL)}\cite{patil2025bfcl} isolates structured function-calling accuracy, focusing on correctness of names, arguments, and outputs.  
Safety-oriented benchmarks such as \textbf{ST-WebAgentBench}\cite{levy2024stwebagentbench} probe robustness under adversarial or policy-sensitive settings.  
Recent extensions like \textbf{BenchAgents}\cite{butt2024benchagents} and \textbf{Benchmark Self-Evolving}\cite{wang2024benchmark} explore automated benchmark generation through agent interactions, complementing our focus on automated data synthesis.

\subsection{Agentic Dataset Construction and Synthetic Data Generation}

Training effective tool-using agents requires rich datasets of tool calls, trajectories, and dialogues.  
\textbf{Mind2Web}\cite{deng2023mind2web} relies on human-annotated trajectories, while \textbf{AgentTrek}\cite{xu2025agenttrek} converts web tutorials into executable trajectories.  
More recently, a growing body of work emphasizes \emph{LLM-only synthetic data generation}.  
\textbf{Self-Instruct}\cite{wang2022selfinstruct} pioneered LLM-only instruction synthesis, while \textbf{Magpie}\cite{xu2024magpie} extended this to alignment data generation from scratch.  
\textbf{UltraFeedback}\cite{cui2024ultrafeedback} scaled preference data collection via AI feedback, whereas \textbf{Arena Learning}\cite{luo2024arena} proposed a simulation-based flywheel for iterative post-training data.  
\textbf{Ge et al. (2024)} demonstrated extreme-scale persona-based synthetic generation, while \textbf{Lee et al. (2024)} introduced iterative LLM-to-LLM refinement pipelines.  
These works highlight scalability but often lack schema-constrained validation. In contrast, our framework enforces strict execution semantics, producing verifiable, machine-checkable agentic records.

\subsection{Multi-Agent Frameworks and Conversational Synthesis}

Several frameworks adopt multi-agent or self-play paradigms for generating data.  
\textbf{AutoGen}\cite{wu2023autogen} enables multi-agent conversations for synthetic supervision, while \textbf{CAMEL}\cite{li2023camel} introduces communicative agents for role-played self-play.  
\textbf{AgentInstruct}\cite{mitra2024agentinstruct} frames dataset construction as generative teaching via agentic flows.  
Related conversational datasets such as \textbf{Baize}\cite{xu2023baize}, \textbf{Zephyr}\cite{tunstall2023zephyr}, \textbf{HelpSteer}\cite{wang2023helpsteer}, and \textbf{LAB}\cite{sudalairaj2024lab} demonstrate alignment through synthetic dialogues, but without the structured execution guarantees provided by our pipelines.

\subsection{Tool Use, Function Calling, and Execution Traces}

A complementary line of research emphasizes explicit tool-use supervision.  
\textbf{ToolCoder}\cite{zhang2023toolcoder} trains models to call APIs via search tools, highlighting the importance of function schemas.  
Reasoning-trace datasets such as \textbf{Program-of-Thoughts}\cite{chen2022program}, \textbf{Orca}\cite{mukherjee2023orca}, \textbf{STAR}\cite{zelikman2022star}, and \textbf{DeepSeekMath}\cite{shao2024deepseekmath} provide structured intermediate reasoning, but do not enforce end-to-end execution of tool calls.  
\textbf{Schluntz and Zhang (2024)} provide practical design principles for building effective agents, aligning with our emphasis on constrained, schema-valid pipelines.

\subsection{Data Quality, Risks, and Governance}

Synthetic data introduces risks of degeneration and bias.  
\textbf{The curse of recursion}\cite{shumailov2023curse} warns of performance collapse when models are trained on self-generated data.  
\textbf{Gudibande et al. (2023)} caution against imitating proprietary LLMs, underscoring the need for modular and diverse generation strategies.  
Empirical studies highlight the importance of quality control: \textbf{Budach et al. (2022)} analyze data quality effects, while surveys on data-centric AI\cite{zha2023datacentric} and best practices for synthetic data\cite{liu2024bestpractices} codify standards for governance.  
Our validators, schema checks, and judge modules directly address these challenges, ensuring reliability and reproducibility in generated datasets.

In summary, prior work has either (a) relied on controlled environments with runtime execution\cite{zhou2024webarena, chezelles2024browsergym}, (b) depended on human-authored demonstrations\cite{deng2023mind2web}, or (c) generated synthetic instructions without tool-specific validation\cite{wang2022selfinstruct, xu2024magpie}.  
Our framework distinguishes itself by integrating these directions: it generates complete domain-to-data pipelines (task definitions, tools, policies, and dialogues), composes them via DAG (Directed Acyclic Graph) abstractions, and validates them with strict structural and semantic checks. This enables scalable, schema-constrained creation of high-fidelity tool-use datasets.

\section{Methodology}
This framework operationalizes the premise that autonomous agents require training data beyond free-form text—specifically, data that captures task intent, tool specifications, execution traces, and dialogue-level interactions in a machine-checkable format.

All experimentation and dataset construction in this work were conducted using the \textbf{SyGra}\footnote{\url{https://github.com/ServiceNow/SyGra}} framework~\cite{pradhan2025sygra}, which serves as the underlying infrastructure for synthetic data orchestration within \textsc{FABRIC}. Within our methodology, SyGra is used as the substrate for defining, executing, and managing the pipeline graphs that power the four data-generation modules described below, ensuring reproducibility and structural consistency across all experimental settings.

Importantly, these data-generation pipelines are designed to be modular rather than interdependent: each pipeline can be executed independently to produce its corresponding supervision signal, or they may be composed sequentially to form an end-to-end data generation workflow. The four pipelines are:

\begin{itemize}
\item \ref{sec:pipeline1} \textbf{RecordSynth}: constructs complete multitask, single-turn agentic records from minimal task seeds, including tool inventories, pseudocode policies, and execution traces;
\item \ref{sec:DAGFirstGeneration} \textbf{DAGFirstGeneration}: decomposes execution DAGs into atomic supervision units—utterances, tool calls, and expected outputs—creating single-task, single-turn agentic data suitable for fine-grained training and evaluation (inspired by BFCL-style benchmarks);
\item \ref{sec:MultiTurnDialogueSynth} \textbf{MultiTurnDialogueSynth}: generates validator-backed multi-turn dialogues, either directly from task prompts or grounded in \textbf{RecordSynth} artifacts;
\item \ref{sec:agentic-record-rollout} \textbf{AgenticRecordRollout}: produces structurally diverse and counterfactual task-graph variants to expand coverage and test robustness.
\end{itemize}

By design, RecordSynth enables full trajectory construction, DAGFirstGeneration isolates atomic function-calling behavior, MultiTurnDialogueSynth focuses on naturalistic multi-turn exchanges, and AgenticRecordRollout enriches structural diversity. Depending on the requirements of a training or evaluation setup, practitioners may invoke these pipelines individually to obtain a specific data type or jointly to build comprehensive, multi-granularity datasets.

The following subsections detail each pipeline, describing their respective inputs, outputs, and generation stages.

\subsection{RecordSynth: Multitask Agentic Data from Scratch}
\label{sec:pipeline1}

The \textbf{RecordSynth} pipeline constructs complete agentic data records from minimal input: an \emph{Enterprise Domain} and an associated \emph{Use Case}. Unlike systems that extract trajectories from observed behavior or demonstrations, this pipeline synthesizes all components of an agentic record \emph{de novo}, relying solely on large language model generations constrained by a unified schema. These records serve as foundational building blocks for both training and evaluation, and provide the source of truth for the derivations in Pipelines 2 and 3.

The output of this pipeline is highly structured: it includes an agentic scenario, a function inventory with formal specifications, a pseudo-code policy sketch, task variants, a grounded user scenario, and an execution trace annotated with sequential and parallel tool invocations. All generated fields conform to strict formatting and consistency constraints to ensure machine-checkability and downstream usability.

\begin{figure}[t]
\centering
\includegraphics[width=0.65\linewidth,trim=0 0 0 30,clip]{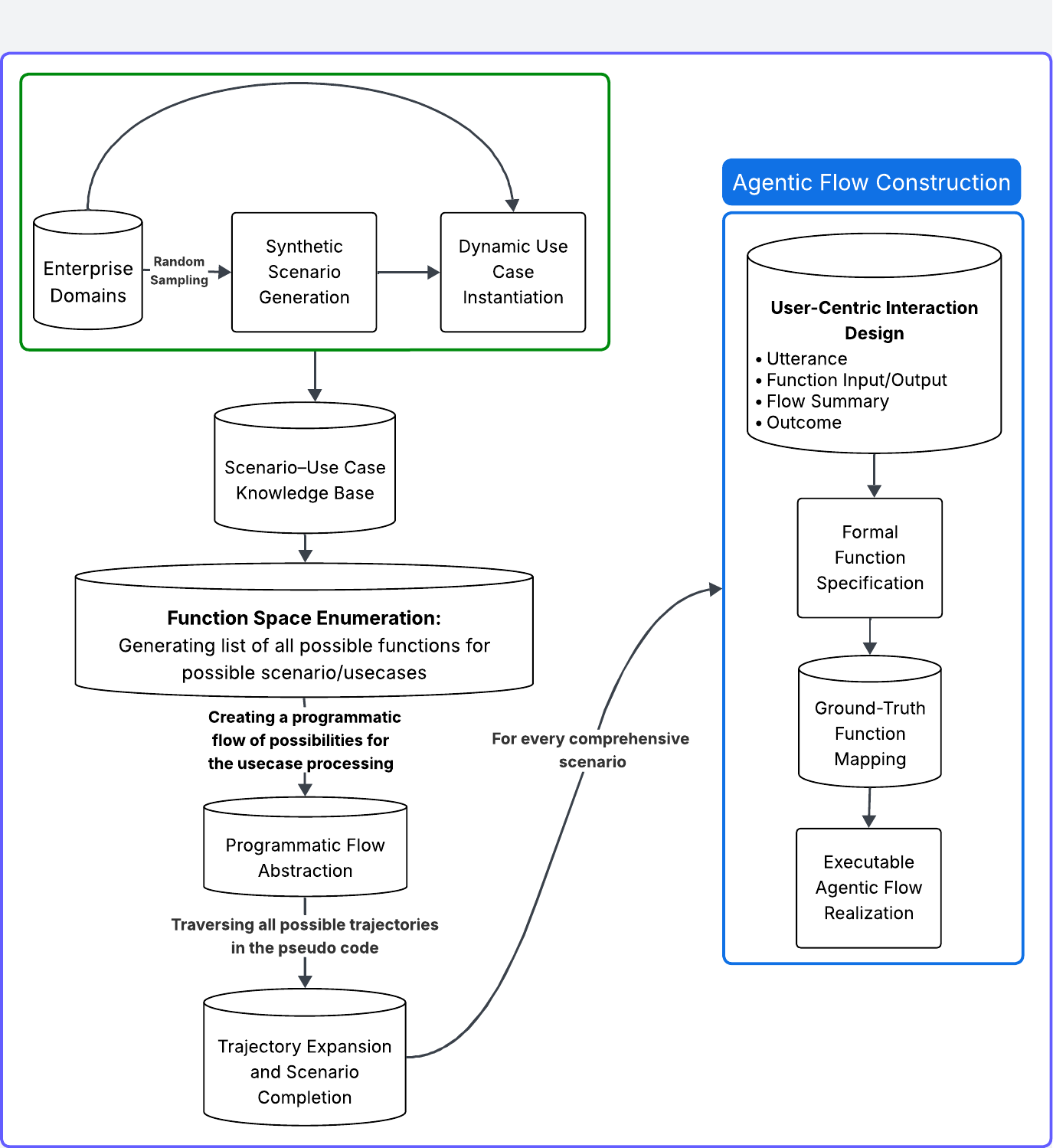}
\caption{End-to-end pipeline for generating multitask agentic data, from enterprise-level scenario construction to executable agentic flows with ground-truth functions.}
\label{fig:pipeline1}
\end{figure}

\subsubsection{Inputs and Outputs}
\label{subsec:p1-io}

\paragraph{Inputs.}
The pipeline accepts:
\begin{itemize}
  \item An \textbf{Enterprise Domain} label: \texttt{\{\{domain\}\}}
  \item A natural language use case: \texttt{\{\{use\_case\}\}}
\end{itemize}

\paragraph{Outputs.}
The pipeline emits a structured artifact consisting of the following components:

\begin{itemize}
  \item \textbf{Agentic Scenario:} A single-paragraph task description in plain English, with no preamble or formatting.
  
  \item \textbf{Functions List (Function Space Enumeration):} A Python list, \texttt{functions\_list = [\,\dots\,]}, where each entry is formatted as: \\
  \texttt{'function\_name(arg1, ...) -> return\_type \# Description'} \\
  with \texttt{return\_type} $\in$ \texttt{\{str, bool, list, dict, None\}}.

  \item \textbf{Pseudo Code (Programmatic Flow Abstraction):} Executable Python code implementing the agent policy, using only functions declared in \texttt{functions\_list}. The code is formatted without preamble or explanatory comments.

  \item \textbf{Possible Scenarios:} A Python \texttt{list[str]} of \texttt{\{\{num\}\}} alternative use cases, each derived from the same tools and logic.

  \item \textbf{User Scenario Pack (User-Centric Interaction Design):} A JSON object of the form:
\begin{verbatim}
{
  "utterance": "...",
  "function_inputs_outputs": [{"function": "...", "input": ..., "output": ...}, ...],
  "flow_summary": "...",
  "outcome": "..."
}
\end{verbatim}
  which must be syntactically valid and directly loadable via \texttt{json.loads()}.

  \item \textbf{Function Descriptions (Formal Function Specification):} A JSON list specifying each function's:
  \texttt{"name"}, \texttt{"properties"} (argument types, defaults, descriptions), \texttt{"required"}, \texttt{"type"} = \texttt{"object"}, \texttt{"description"}, and \texttt{"output"} with type and description.

  \item \textbf{Agentic Execution (Executable Agentic Flow Realization):} A JSON list of step objects, each structured as:
\begin{verbatim}
{
  "step_number": ...,
  "functions_executed": [...],
  "is_parallel": true/false,
  "description": "...",
  "input": [...],
  "output": [...]
}
\end{verbatim}
  Multi-function steps are marked with \texttt{is\_parallel=true}. Inputs and outputs are aligned positionally with the functions listed in \texttt{functions\_executed}.
\end{itemize}

\subsubsection{Generation Stages}
\label{subsec:agentic-record-synth-stages}

The generation process is structured into eight sequential steps, each responsible for producing a schema-conformant component of the agentic record. Each step also encodes a design rationale explaining \emph{why} it is necessary for coverage, completeness, or realism.

\begin{itemize}
  \item \textbf{Step 1: Create Scenarios.}  
  Given an enterprise domain, generate \texttt{\{\{num\}\}} plausible and diverse use cases as a Python \texttt{list[str]}. Each item must be a plain-English string with no preamble or formatting.  
  This step ensures breadth and variability. By sampling multiple use cases within an enterprise domain, the framework establishes the outer boundary of possible agentic behaviors before narrowing toward specific flows.

  \item \textbf{Step 2: Create Agentic Scenario.}  
  For a selected use case, synthesize a concise, single-paragraph description in plain English that describes the real-time context in which the agent operates.  
  The goal is realism and grounding. Abstract tasks are expanded into operational contexts to help the model capture situational cues and implicit dependencies—essential for naturalistic trajectory generation.

  \item \textbf{Step 3: Create Functions List (Function Space Enumeration).}  
  Emit a Python-style list assignment \texttt{functions\_list = [\,\dots\,]}, where each entry has the form \texttt{'function\_name(arg1, ...) -> return\_type \# Description'}. The return type must belong to \texttt{\{str, bool, list, dict, None\}}. This list acts as the foundation of tool availability. Enumerating all functions ensures dependency completeness and prevents gaps in the agent’s action space during pseudo-code synthesis.

  \item \textbf{Step 4: Create Pseudo Code (Programmatic Flow Abstraction).}  
  Generate executable (pydantic-compatible) Python code that implements the agent’s decision logic using only the functions defined in \texttt{functions\_list}. The output must contain no explanatory comments or preamble. Pseudo code serves as the structural backbone of the agentic record. It ensures coverage by explicitly linking all dependencies and defining logical flow paths that the model can later expand into real trajectories.

  \item \textbf{Step 5: Create Possible Scenarios.}  
  Derive \texttt{\{\{num\}\}} alternative use cases based on the same function inventory and policy logic. These are emitted as a Python \texttt{list[str]} with plain-English descriptions. Expanding from one to many related scenarios introduces flexibility and realism. It captures the inherent variability in enterprise tasks and helps the model generalize without bias to a single flow.

  \item \textbf{Step 6: Create User Scenario Pack (User-Centric Interaction Design).}  
  Construct a JSON object with the structure:
    \begin{listing}[!htpb]
    \begin{center}
    \begin{minted}[
        gobble=4,
        fontsize=\scriptsize,
        breaklines=true,
    ]{json}
                        {
                          "utterance": "...",
                          "function_inputs_outputs": [
                            {"function": "...", "input": ..., "output": ...},
                            ...
                          ],
                          "flow_summary": "...",
                          "outcome": "..."
                        }
    \end{minted}
    \end{center}
    \end{listing}
  This object represents the user-facing trajectory and must be directly loadable via \texttt{json.loads()}. By aligning utterances and function calls, this step grounds formal computation in natural dialogue. It bridges human intention and machine execution, making supervision data more interpretable.

  \item \textbf{Step 7: Create Function Descriptions (Formal Function Specification).}  
  Convert each entry in \texttt{functions\_list} into a detailed JSON schema with fields: \texttt{"name"}, \texttt{"properties"} (typed arguments with defaults and descriptions), \texttt{"required"}, \texttt{"type"} = \texttt{"object"}, \texttt{"description"}, and \texttt{"output"}. This ensures precision and reproducibility. Each function becomes a typed, verifiable object definition, reducing ambiguity and enabling the data to be programmatically validated and re-used.

  \item \textbf{Step 8: Create Agentic Execution (Executable Agentic Flow Realization).}  
  Produce a JSON list of ordered steps, each capturing the functions executed, execution mode (\texttt{is\_parallel}), inputs, outputs, and a description. Each step has the following structure:
    \begin{listing}[!htpb]
    \begin{center}
    \begin{minted}[
        gobble=4,
        fontsize=\scriptsize,
        breaklines=true,
    ]{json}
                        {
                          "step_number": ...,
                          "functions_executed": [...],
                          "is_parallel": true/false,
                          "description": "...",
                          "input": [...],
                          "output": [...]
                        }
    \end{minted}
    \end{center}
    \end{listing}
  Functions that can be executed concurrently are grouped under a single step with \texttt{is\_parallel=true}. The \texttt{input} and \texttt{output} lists must align positionally with the functions listed. This final stage formalizes the full agentic trajectory. It captures execution diversity (sequential vs. parallel) and ensures machine-readability for both replay and evaluation.
\end{itemize}

This pipeline establishes the foundation for scalable, LLM-based synthesis of structured agentic datasets. Its outputs are both human-readable and machine-checkable, enabling deterministic parsing, structural validation, and seamless integration into downstream derivations. These records serve as canonical sources for generating single-turn and multi-turn variants used in tool-use training and evaluation settings.

\subsection{DAGFirstGeneration: Agentic Data with DAG First Approach}
\label{sec:DAGFirstGeneration}

This pipeline targets the generation of fine-grained, tool-level supervision derived from the execution DAGs constructed in \ref{sec:pipeline1}. Given a complete record from \textbf{RecordSynth}, each node of the execution DAG---representing a single tool invocation with defined arguments and outputs---can be isolated into an atomic supervision unit. The resulting data takes the form of structured triples: \texttt{(user utterance, tool call, expected output)}.

\begin{figure}[H]
\centering
\includegraphics[width=0.95\linewidth,trim=0 0 0 30,clip]{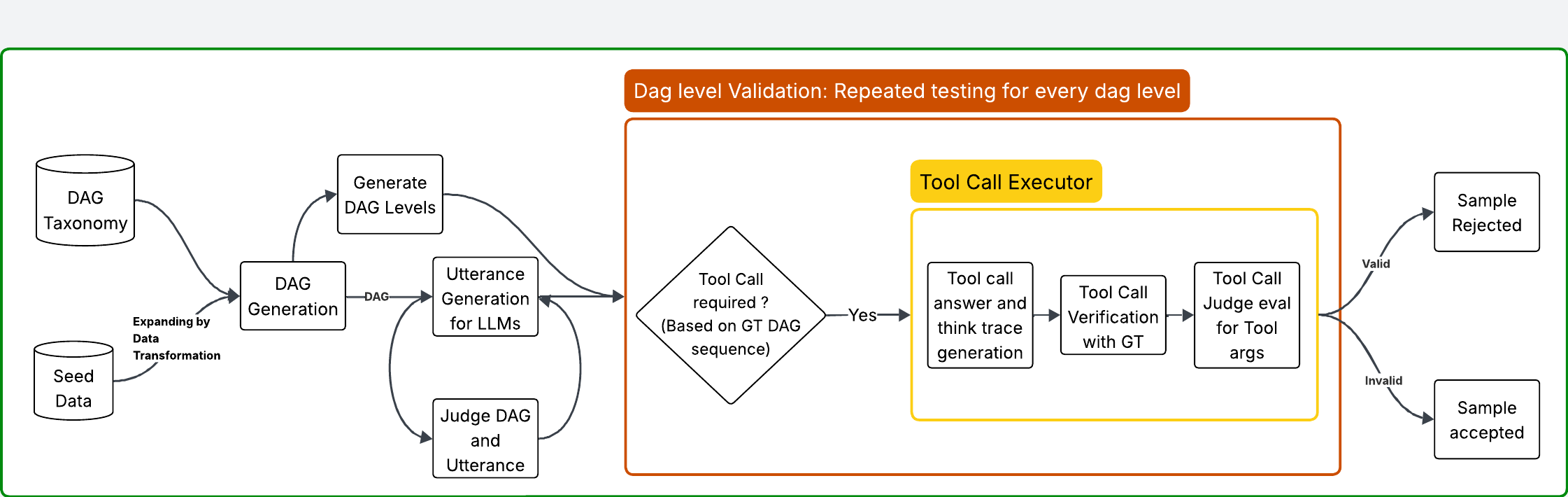}
\caption{Agentic Data with DAG First Approach}
\label{fig:pipeline3-bfcl}
\end{figure}

By decomposing end-to-end traces into these minimal units, the pipeline enables precise supervision for grounding natural language into tool arguments and verifying functional correctness. Each triple is schema-validated, ensuring that argument types, dependencies, and outputs remain consistent with the original DAG. This approach captures the local semantics of tool use while maintaining alignment with the global execution trace.

Although related to function-calling benchmarks such as the Berkeley Function-Calling Leaderboard (BFCL), the emphasis here is broader: the pipeline systematically produces atomic tool-call data at scale from arbitrary DAGs, rather than being restricted to any one benchmark format. In practice, these outputs support both targeted fine-tuning and evaluation of LLM agents on structured tool use.

\subsubsection{Generation and Validation Stages}

\begin{itemize}

\item \textbf{Stage 1: DAG Seeding and Expansion.}  
A core taxonomy of Directed Acyclic Graphs (DAGs) serves as the structural backbone for function-calling strategies (e.g., linear chains, diamonds, fan-in, conditionals). Each DAG represents an archetypal multi-tool reasoning flow, instantiated from agentic plans generated in Pipeline~1. DAG variants \ref{fig:dag} are expanded using automated rewrites (e.g., node substitutions, edge permutations) to increase diversity.

\item \textbf{Stage 2: DAG-Level Decomposition.}  
Each DAG is decomposed into per-node atomic units. For every tool node:
\begin{itemize}
    \item All tool input requirements are computed via topological traversal over upstream outputs and user inputs.
    \item Ground truth function name and arguments are retrieved from the execution trace.
    \item The node is transformed into a BFCL-style triple: 
    \begin{quote}
        \texttt{(user utterance, tool call, expected output)}
    \end{quote}
    where the utterance is scoped only to the inputs available up to that node.
\end{itemize}

\begin{figure}[!htbp]
\centering
\includegraphics[width=0.95\linewidth,trim=0 0 0 30,clip]{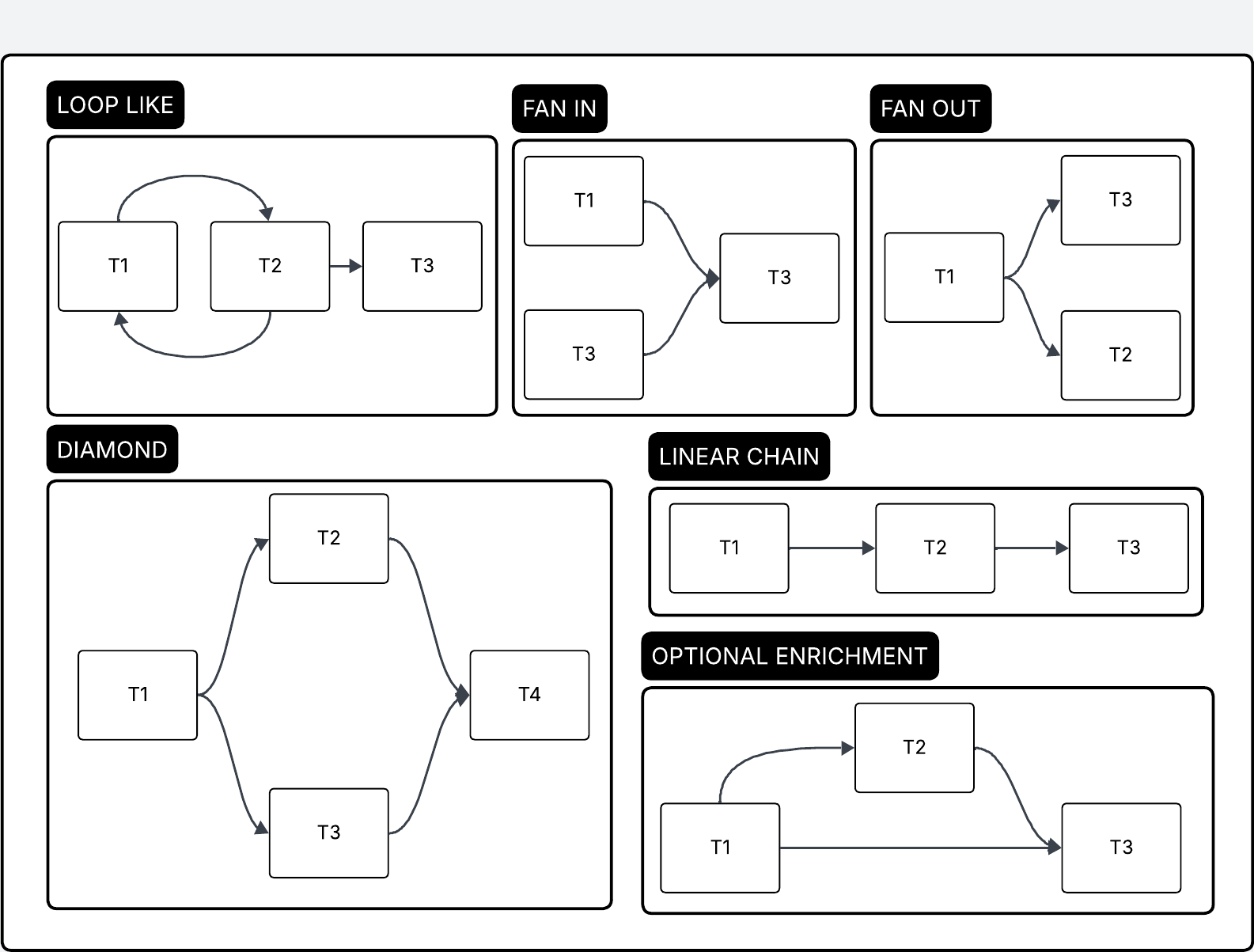}
\caption{Illustration of different DAG structures used in task orchestration. Variants include (i) Fan-out/Fan-in, where a single task branches into parallel tasks before converging; (ii) Linear chain, with tasks executed sequentially; (iii) Loop-like pattern, representing iterative dependencies; (iv) Diamond structure, where a task splits into multiple dependent paths that later rejoin; and (v) Optional enrichment, where supplementary tasks may or may not be executed depending on context. These patterns capture common workflow designs in agentic LLM systems.}
\label{fig:dag}
\end{figure}

\item \textbf{Stage 3: User Utterance Generation.}  
Each atomic tool call is turned into a natural user instruction via prompt-based generation:
\begin{itemize}
    \item For \textbf{direct style}, all required arguments are embedded explicitly in the utterance.
    \item For \textbf{indirect style}, some arguments are implied via common-sense or contextual inference.
    \item Prompts are conditioned on function schema, input types, and any prior tool outputs available at that node.
\end{itemize}
These utterances are evaluated for specificity, fluency, and uniqueness—ensuring they map cleanly and exclusively to the intended function call.

\item \textbf{Stage 4: Tool Call Simulation and Trace Generation.}  
This stage simulates how an LLM agent would respond to the utterance:
\begin{itemize}
    \item A \textbf{think trace} is generated to document the LLM’s inner reasoning: how each input was grounded from prior context or user instruction.
    \item A structured \textbf{tool call JSON} is produced, matching the required function and argument format.
\end{itemize}
Importantly, the model is prohibited from referencing DAG structure, tool names, or internal schemas in its reasoning—only the input-output grounding is permitted.

\item \textbf{Stage 5: Tool Call Verification.}  
The generated tool call is automatically validated against the DAG ground truth and JSON schema:
\begin{itemize}
    \item Function name must match the correct node.
    \item All required argument fields must be present, well-typed, and semantically valid.
    \item No hallucinated or extraneous fields are allowed.
\end{itemize}
This automated verification ensures that only structurally executable and schema-valid samples pass through.

\item \textbf{Stage 6: Judge Evaluation and Scoring.}  
Finally, each \texttt{(utterance, tool call)} pair is reviewed using a judging module, which scores on:
\begin{itemize}
    \item Instruction clarity and specificity;
    \item Argument sufficiency and grounding;
    \item Alignment with argument formatting style (direct vs. indirect);
    \item Naturalness and user-centric phrasing.
\end{itemize}
Samples rated below threshold are filtered out; accepted ones populate the final BFCL-ready dataset.

\end{itemize}

In summary, our methodology establishes a principled framework for automated agentic data synthesis. By chaining three pipelines with distinct granularity levels, we achieve:

\begin{itemize}
    \item \textbf{Compositionality:} Data is generated across multiple abstraction levels (record $\rightarrow$ dialogue $\rightarrow$ atomic function call), enabling flexible reuse.
    \item \textbf{Validation:} Each stage includes schema enforcement, type checking, and judge-based filtering to mitigate errors and hallucinations common in LLM-only generation.
    \item \textbf{Scalability:} The pipelines operate without human-in-the-loop supervision, supporting massive dataset creation across domains, tools, and interaction styles.
    \item \textbf{Benchmark compatibility:} Outputs are aligned with both multi-turn interaction benchmarks and BFCL-style function-calling leaderboards, ensuring direct evaluation relevance.
\end{itemize}

This modular pipeline design not only supports scalable dataset generation but also provides a blueprint for systematically studying how LLMs learn tool-use reasoning at different levels of granularity.

\subsection{MultiTurnDialogueSynth: Multi-Turn Agentic Dialogue Generation}
\label{sec:MultiTurnDialogueSynth}

This pipeline constructs an $N$-turn structured conversation between a user proxy and an LM-based agent. The conversation unfolds as a series of goal-directed exchanges involving function calls, tool responses, and interaction-level validation. Unlike \textsc{AgenticRecordSynth}, which generates a static execution trace, this component simulates a dynamic turn-by-turn dialogue, thereby modeling more realistic usage scenarios for agentic LMs. All interactions are grounded in the artifacts derived from the seed record produced by \textsc{AgenticRecordSynth}.

\begin{figure}[htbp]
\centering
\includegraphics[width=\linewidth,trim=0 0 0 30,clip]{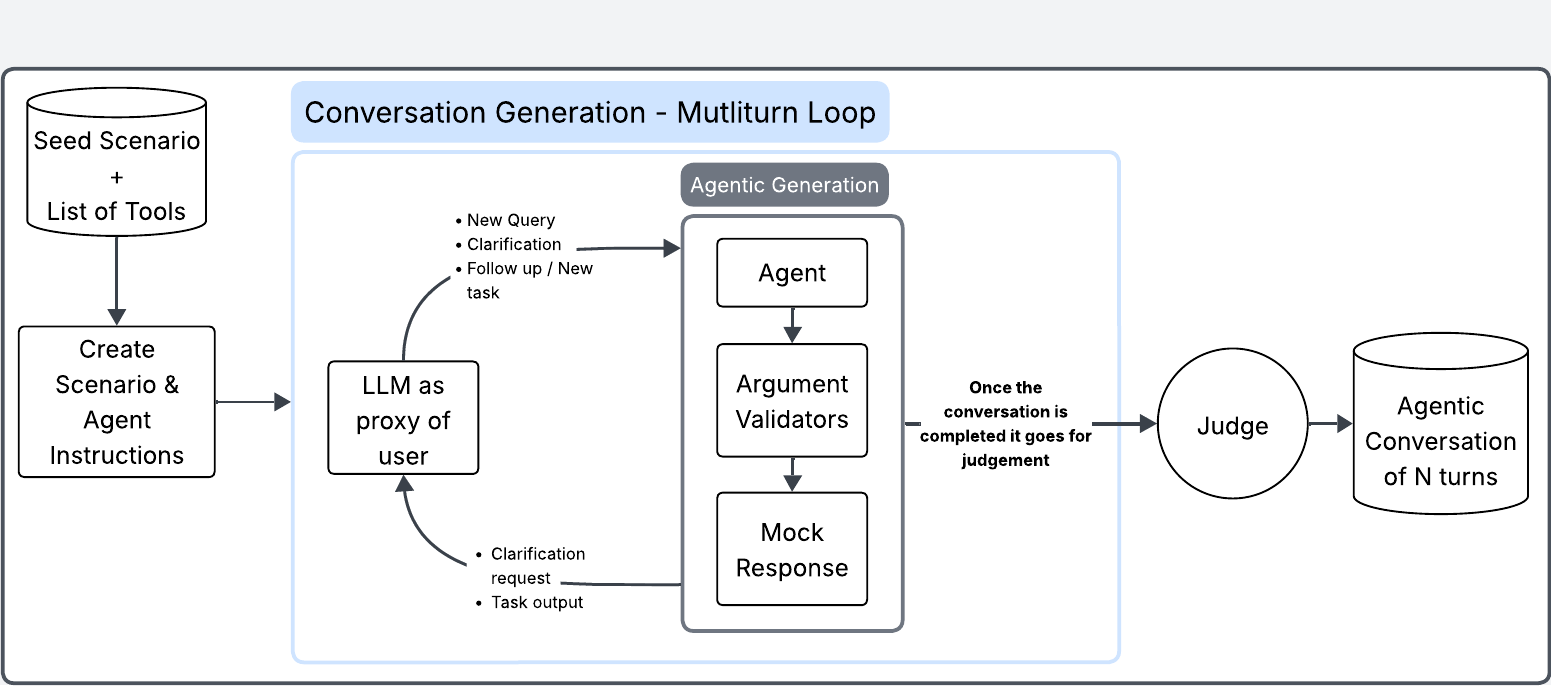}
\caption{Turn-level interaction flow in \textsc{MultiTurnDialogueSynth}: each step consists of user input generation, agent tool invocation, argument validation, mocked response generation, and judge feedback.}
\label{fig:agentic-dialogue-synth}
\end{figure}

\subsubsection{Inputs and Outputs}
\label{subsec:dialogue-io}

\paragraph{Inputs.}
This component takes as input a seed record generated by the pipeline \ref{subsec:agentic-record-synth-stages}, specifically:
\begin{itemize}
  \item \texttt{agentic\_scenario} — the narrative context for the task;
  \item \texttt{functions\_list} — the available tool signatures;
  \item \texttt{function\_description} — structured specifications for each tool;
  \item \texttt{pseudo\_code} — the high-level agent policy;
  \item a selected \texttt{possible\_scenario} — the specific task variant for the dialogue.
\end{itemize}

\paragraph{Output.}
An \emph{agentic conversation} of $N$ validated turns, represented as a structured transcript containing:
\begin{itemize}
  \item natural language messages from both user and agent;
  \item tool calls with arguments;
  \item validated responses based on mock specifications;
  \item turn-level judgments governing progression or termination.
\end{itemize}

\subsubsection{Conversation Generation Process}
\label{subsec:dialogue-generation}

The dialogue is constructed via the following two-stage procedure:

\begin{itemize}
  \item \textbf{Step 1: Initialize Scenario and Agent Instructions.}  
  Based on the \texttt{agentic\_scenario}, \texttt{functions\_list}, and \texttt{pseudo\_code}, the system initializes contextual instructions for both the user proxy and the agent. These instructions define allowable actions, grounded tools, and the underlying policy logic.

  \item \textbf{Step 2: Multi-Turn Simulation Loop.}  
  The conversation proceeds iteratively through a turn-generation loop. At each iteration:
  \begin{enumerate}
    \item The \textbf{user proxy} (LLM) generates a new input (query, clarification, or follow-up) consistent with the seeded task. Refer to prompt templates (\ref{lst:pp3_generate_user1}, \ref{lst:pp3_generate_user2}).
    \item The \textbf{agent} responds by issuing tool calls defined in \texttt{function\_description}, using correct function names, arguments, and output types.
    \item An \textbf{argument validator} checks for type correctness and schema alignment.
    \item A \textbf{mock response generator} returns tool outputs consistent with the function specification and prior example I/O, refer to prompts (\ref{lst:pp3_mock_response}). Mock response simulates realistic failure cases which agent must gracefully fix in the conversation hence convering diverse scenarios.
    \item A \textbf{judge module} evaluates the turn (user input + agent response + mock output) and returns a signal: \texttt{accept}, \texttt{revise}, or \texttt{terminate}.
  \end{enumerate}
  Rejected turns are revised and resubmitted until accepted. The loop terminates either on task completion or when the judge emits a stop signal.
\end{itemize}

\subsubsection{Turn-Level Structure and Representation}
\label{subsec:dialogue-turn-format}

Each conversation is stored as an ordered list of structured turn objects, where each turn contains:

\begin{itemize}
  \item \texttt{role}: either \texttt{"user"} or \texttt{"assistant"};
  \item \texttt{content}: the natural language utterance (no preamble or markdown);
  \item \texttt{tool\_calls} (assistant only): zero or more validated calls in the form \texttt{name(args)}, consistent with \texttt{function\_description};
  \item \texttt{tool\_results} (assistant only): mocked return values aligned to the issued calls;
  \item \texttt{judge\_decision}: one of \texttt{"accept"}, \texttt{"revise"}, or \texttt{"terminate"}.
\end{itemize}

The final conversation constitutes a validated $N$-turn trajectory that models realistic agentic interaction grounded in structured tool-use and policy-consistent behavior. These conversations may be used for multi-turn fine-tuning, preference modeling, or robustness evaluation of tool-augmented agents.

\subsection{AgenticRecordRollout: Controlled Deployment of Agentic Records}
\label{sec:agentic-record-rollout}

\begin{figure}[!htbp]
\centering
\includegraphics[width=0.75\linewidth,trim=0 0 0 22,clip]{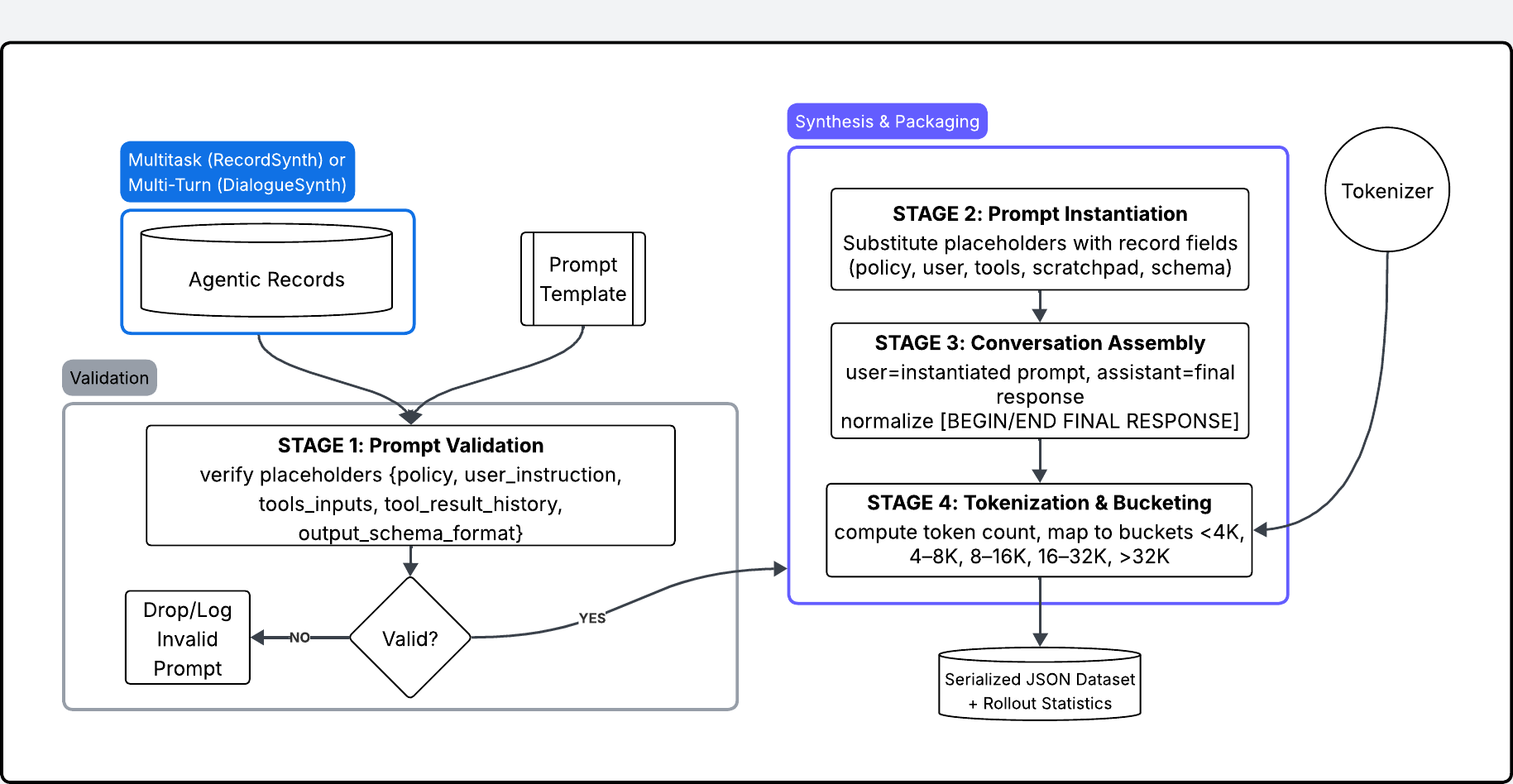}
\caption{Agentic Record Rollout Pipeline. Inputs (agentic records, prompt templates, tokenizer) flow through validation, instantiation, conversation assembly, and tokenization/bucketing to produce serialized JSON datasets and rollout statistics.}
\label{fig:agentic-rollout}
\end{figure}

The fourth pipeline operationalizes the \emph{rollout} of agentic records into training-ready (SFT) datasets. Whereas the previous pipelines focus on data synthesis at varying granularities (end-to-end trajectories, multi-turn dialogues), this pipeline ensures that generated records are properly validated, bucketed, and serialized into chat format ready to consumed for SFT. Importantly, the rollout stage is agnostic to whether the source artifacts are derived from the multitask pipeline (Sec.~\ref{sec:pipeline1}) or the multi-turn pipeline (Sec.~\ref{sec:MultiTurnDialogueSynth}), enabling a unified procedure for downstream usage.

\subsubsection{Inputs and Outputs}
\label{subsec:rollout-io}

\paragraph{Inputs.}
\begin{itemize}
  \item \texttt{agentic\_records}: Structured records generated by any of the upstream pipelines (RecordSynth, DialogueSynth).
  \item \texttt{prompt\_templates}: A JSON collection of base prompts parameterized with placeholders for \texttt{policy}, \texttt{user\_instruction}, \texttt{tools\_inputs}, \texttt{tool\_result\_history}, and \texttt{output\_schema\_format}.
  \item \texttt{tokenizer}: A pretrained tokenizer (e.g., \texttt{Mistral-Nemo-Instruct-2407}) for computing token lengths and assigning bucket labels.
\end{itemize}

\paragraph{Outputs.}
\begin{itemize}
  \item A validated and serialized JSON dataset in which each entry contains:
  \begin{enumerate}
    \item A unique identifier (\texttt{uuid});
    \item A two-turn conversation consisting of user prompt and assistant response;
    \item A taxonomy annotation (\texttt{Agentic}, token length, token bucket);
    \item Metadata fields and tags for downstream filtering (e.g., \texttt{agentic}, \texttt{reasoning}, \texttt{thinking});
    \item Model annotations for evaluation alignment.
  \end{enumerate}
  \item Statistics on bucket distributions, rejection counts, and metadata coverage.
\end{itemize}

\subsubsection{Rollout Procedure}
\label{subsec:rollout-procedure}

The rollout process is organized into four technical stages:

\begin{itemize}
  \item \textbf{Stage 1: Prompt Validation and Selection.}  
  Candidate prompts are filtered using a schema validator that ensures the presence of all required placeholders. Invalid prompts are discarded prior to expansion.

  \item \textbf{Stage 2: Prompt Instantiation.}  
  For each agentic record, a validated prompt template is selected at random. Placeholders are substituted with record-specific content: user utterances, tool input histories, scratchpads, and output schemas.

  \item \textbf{Stage 3: Conversation Assembly.}  
  A synthetic two-turn dialogue is constructed, pairing the instantiated prompt (user role) with the model’s final response (assistant role). Special tokens such as \texttt{[BEGIN FINAL RESPONSE]} and \texttt{[END FINAL RESPONSE]} are normalized to enforce consistency.
\end{itemize}

\section{Conclusion}
\label{sec:conclusion}

We introduce a unified, modular framework for generating structured, executable, and validated tool-use data from large language models. By decomposing the generation process into DAG-based agentic planning (Pipeline~1), multi-step instruction synthesis (Pipeline~2), and atomic BFCL-style derivations (Pipeline~3), our approach enables high-fidelity benchmarking and training across multiple abstraction levels of tool-augmented reasoning.

The core insights of this framework lie in:
\begin{itemize}
    \item Leveraging declarative DAG templates to anchor tool dependencies;
    \item Using execution traces as supervision signals for grounded instruction generation;
    \item Embedding structured validation and judge scoring to enforce output correctness.
\end{itemize}

Together, these elements bridge the gap between free-form language generation and robust, schema-aligned tool calling—paving the way for scalable, reproducible evaluation of tool-augmented LLM agents.

We hope this framework serves as a foundation for future work in tool-learning, DAG supervision, and compositional reasoning. Extensions could include support for conditional branching, multi-agent settings, long-term memory traces, or continual learning across toolchains.

\section{Limitations and Considerations}
\label{sec:limitations}

Despite its modularity and empirical robustness, our proposed framework presents several limitations that merit discussion:

\subsection{Threats to Validity}
While the multi-pipeline design enforces structure and validation at each stage, the overall correctness of generated samples ultimately depends on upstream assumptions:
\begin{itemize}
    \item \textbf{DAG fidelity:} Errors in DAG instantiation or tool interdependencies may propagate into lower-level tool-call supervision, leading to incorrect or misleading ground truths.
    \item \textbf{Prompt sensitivity:} Output quality is highly dependent on the design of system and user prompts. Subtle phrasing changes can lead to cascading effects in argument inference or reasoning quality.
\end{itemize}

\subsection{Bias and Safety Considerations}
Generated utterances and tool call traces are subject to the biases of the underlying LLM:
\begin{itemize}
    \item \textbf{Data exposure:} Hallucinated or unsafe inputs may appear if the model has memorized unsafe tool signatures or real-world values from pretraining.
    \item \textbf{Overfitting to prompt patterns:} Repeated use of fixed prompt templates may encourage unnatural language patterns or memorization of call schemas.
    \item \textbf{Safety-critical misuse:} While our framework targets abstract tool-use benchmarking, any downstream application in real-world agent deployments must incorporate safety-aware tool schemas and rejection filtering.
\end{itemize}

\subsection{Risk of Model Collapse}
As LLMs are increasingly trained on synthetic datasets, there is a long-term risk of \textit{synthetic signal amplification}:
\begin{itemize}
    \item \textbf{Degeneration through self-distillation:} Repeated exposure to simplified traces or schema-conforming calls may flatten the model’s reasoning diversity.
    \item \textbf{Loss of grounding generality:} If training focuses exclusively on DAG-extracted atomic calls, models may fail to generalize to novel tool compositions or out-of-distribution reasoning tasks.
\end{itemize}
Mitigating these risks will require dataset diversification, adversarial sampling, and continual grounding in real-world tasks.

\subsection{Reproducibility Constraints}
Although our pipelines are modular and defined via templated prompts and execution harnesses, full reproduction requires:
\begin{itemize}
    \item Access to consistent function libraries and schemas.
    \item Stable versions of LLMs with deterministic sampling parameters.
    \item Storage and execution infrastructure capable of multi-stage tracing and filtering.
\end{itemize}
We release configurations and intermediate artifacts to support transparency, but exact replication may require retraining or model re-derivation in future settings.

\bibliographystyle{unsrt}  
\bibliography{template}

\newpage

\appendix

\section{Prompt Templates}
\label{app:prompts}

\begin{listing}[!htpb]
\begin{center}
\begin{minted}[
    gobble=4,
    fontsize=\scriptsize,
    breaklines=true,
]{json}
[
    {
        "domain":"Contract Lifecycle Management (CLM)",
        "scenario":"Procurement-CLM Integration: Automate synchronization between signed contracts and procurement systems to enforce pricing, volume, and payment terms during order fulfillment.",
        "agentic_scenario":"A legal team at a multinational corporation uses an AI-driven agent to monitor contract drafts in real time as they're edited in the CLM platform. The agent employs NLP to analyze clauses, automatically flagging ambiguous or non-compliant terms against jurisdiction-specific laws, internal policy databases, and prior approved contracts. It suggests standardized phrasing for risk areas like payment terms or termination conditions while verifying signatory authority levels, ensuring all amendments are logged and compliance gaps resolved before contracts reach final approval, cutting legal review cycles by 40%.",
        "function_list":"functions_list = [\n    'fetch_contract_draft(contract_id) -> dict # Retrieve the current contract draft from the CLM system.',\n    'analyze_contract_text(text) -> list # Parse the contract text into clauses and key terms.',\n    'retrieve_jurisdictional_requirements(jurisdiction) -> list # Obtain relevant laws based on the contract\u2019s jurisdiction.',\n    'get_internal_policies() -> list # Fetch internal policy documents and compliance rules.',\n    'check_clause_compliance(clause, laws, policies) -> list # Identify non-compliant issues in a clause against laws\/policies.',\n    'compare_with_approved_clauses(clause) -> list # Find matching clauses in prior approved contracts for reference.',\n    'suggest_standardized_terms(issue_type, clause_type) -> str # Propose standardized phrasing for ambiguous or risky terms.',\n    'verify_signatory_authority(signatory_role, contract_scope) -> bool # Confirm signatory has required authority level.',\n    'log_contract_amendment(contract_id, change) -> bool # Record all edits\/amendments made to the contract draft.',\n    'generate_compliance_summary(contract_id) -> dict # Create a structured report of flagged issues and resolutions.',\n    'alert_legal_team(contract_id, flagged_issues) -> None # Notify legal stakeholders about critical compliance gaps.',\n    'monitor_real_time_edits(contract_id) -> None # Track live changes and trigger reanalysis automatically.',\n    'check_overall_compliance(contract_id) -> bool # Validate if all flagged issues are resolved for final approval.',\n    'escalate_delayed_resolution(contract_id) -> bool # Automatically escalate unresolved high-risk clauses nearing deadlines.',\n    'learn_from_resolution_outcomes(review_outcome) -> None # Refine AI logic using feedback from final contract approvals.'\n]",
        "function_description":[
            {
                "name":"fetch_contract_draft",
                "properties":{
                    "contract_id":{
                        "type":"string",
                        "description":"ID of the contract to retrieve the draft for."
                    }
                },
                "required":[
                    "contract_id"
                ],
                "type":"object",
                "description":"Retrieve the current contract draft from the CLM system.",
                "output":{
                    "type":"object",
                    "description":"The contract draft details including text and metadata."
                }
            },
            {
                "name":"analyze_contract_text",
                "properties":{
                    "text":{
                        "type":"string",
                        "description":"The contract text to analyze."
                    }
                },
                "required":[
                    "text"
                ],
                "type":"object",
                "description":"Parse the contract text into clauses and key terms.",
                "output":{
                    "type":"array",
                    "description":"List of clauses parsed from the contract text."
                }
            },

                

\end{minted}
\end{center}
\caption{RecordSynth: Sample Record - Part 1}
\label{lst:RecordSynthSampleRecord1}
\end{listing}

\begin{listing}[!htpb]
\begin{center}
\begin{minted}[
    gobble=4,
    fontsize=\scriptsize,
    breaklines=true,
]{json}
[

            {
                "name":"retrieve_jurisdictional_requirements",
                "properties":{
                    "jurisdiction":{
                        "type":"string",
                        "description":"Jurisdiction to retrieve relevant legal requirements for."
                    }
                },
                "required":[
                    "jurisdiction"
                ],
                "type":"object",
                "description":"Obtain relevant laws based on the contract\u2019s jurisdiction.",
                "output":{
                    "type":"array",
                    "description":"List of legal requirements relevant to the jurisdiction."
                }
            },
            {
                "name":"get_internal_policies",
                "properties":{

                },
                "required":[

                ],
                "type":"object",
                "description":"Fetch internal policy documents and compliance rules.",
                "output":{
                    "type":"array",
                    "description":"List of internal company policy documents and compliance rules."
                }
            },
            {
                "name":"check_clause_compliance",
                "properties":{
                    "clause":{
                        "type":"object",
                        "description":"The clause text and metadata to evaluate."
                    },
                    "laws":{
                        "type":"array",
                        "items":{
                            "type":"string"
                        },
                        "description":"List of jurisdiction-specific legal requirements."
                    },
                    "policies":{
                        "type":"array",
                        "items":{
                            "type":"string"
                        },
                        "description":"List of internal company policy rules."
                    }
                },
                "required":[
                    "clause",
                    "laws",
                    "policies"
                ],
                "type":"object",
                "description":"Identify non-compliant issues in a clause against laws\/policies.",
                "output":{
                    "type":"array",
                    "description":"List of compliance issues found in the clause."
                }
            },
            {
                "name":"compare_with_approved_clauses",
                "properties":{
                    "clause":{
                        "type":"object",
                        "description":"Clause to compare against prior approved contracts."
                    }
                },
                "required":[
                    "clause"
                ],

\end{minted}
\end{center}
\caption{RecordSynth: Sample Record - Part 2}
\label{lst:RecordSynthSampleRecord2}
\end{listing}

\begin{listing}[!htpb]
\begin{center}
\begin{minted}[
    gobble=4,
    fontsize=\scriptsize,
    breaklines=true,
]{json}
                "type":"object",
                "description":"Find matching clauses in prior approved contracts for reference.",
                "output":{
                    "type":"array",
                    "description":"List of matching clauses from approved contracts for reference."
                }
            },
            {
                "name":"suggest_standardized_terms",
                "properties":{
                    "issue_type":{
                        "type":"string",
                        "description":"Type of issue to address (e.g., ambiguity, risk)."
                    },
                    "clause_type":{
                        "type":"string",
                        "description":"Type of clause the issue is located in."
                    }
                },
                "required":[
                    "issue_type",
                    "clause_type"
                ],
                "type":"object",
                "description":"Propose standardized phrasing for ambiguous or risky terms.",
                "output":{
                    "type":"string",
                    "description":"Standardized phrasing suggestion to resolve the issue."
                }
            },
            {
                "name":"verify_signatory_authority",
                "properties":{
                    "signatory_role":{
                        "type":"string",
                        "description":"Role of the person signing the contract."
                    },
                    "contract_scope":{
                        "type":"string",
                        "description":"Scope\/authority level required for the contract."
                    }
                },
                "required":[
                    "signatory_role",
                    "contract_scope"
                ],
                "type":"object",
                "description":"Confirm signatory has required authority level.",
                "output":{
                    "type":"boolean",
                    "description":"True if signatory has appropriate authority; false otherwise."
                }
            },
            {
                "name":"log_contract_amendment",
                "properties":{
                    "contract_id":{
                        "type":"string",
                        "description":"ID of the contract being amended."
                    },
                    "change":{
                        "type":"object",
                        "description":"Amendment details including original text and proposed changes."
                    }
                },
                "required":[
                    "contract_id",
                    "change"
                ],

\end{minted}
\end{center}
\caption{RecordSynth: Sample Record - Part 3}
\label{lst:RecordSynthSampleRecord3}
\end{listing}

\begin{listing}[!htpb]
\begin{center}
\begin{minted}[
    gobble=4,
    fontsize=\scriptsize,
    breaklines=true,
]{json}
            "type":"object",
                "description":"Record all edits\/amendments made to the contract draft.",
                "output":{
                    "type":"boolean",
                    "description":"True if amendment was recorded successfully."
                }
            },
            {
                "name":"generate_compliance_summary",
                "properties":{
                    "contract_id":{
                        "type":"string",
                        "description":"ID of the contract to summarize compliance status for."
                    }
                },
                "required":[
                    "contract_id"
                ],
                "type":"object",
                "description":"Create a structured report of flagged issues and resolutions.",
                "output":{
                    "type":"object",
                    "description":"Structured report of compliance issues and resolutions."
                }
            },
            {
                "name":"alert_legal_team",
                "properties":{
                    "contract_id":{
                        "type":"string",
                        "description":"ID of the contract with critical compliance gaps."
                    },
                    "flagged_issues":{
                        "type":"array",
                        "items":{
                            "type":"string"
                        },
                        "description":"List of critical compliance issues requiring legal review."
                    }
                },
                "required":[
                    "contract_id",
                    "flagged_issues"
                ],
                "type":"object",
                "description":"Notify legal stakeholders about critical compliance gaps.",
                "output":{
                    "type":"null",
                    "description":"Triggers legal team notification for critical issues; returns None."
                }
            },
            {
                "name":"monitor_real_time_edits",
                "properties":{
                    "contract_id":{
                        "type":"string",
                        "description":"Contract ID to monitor for real-time edits."
                    }
                },
                "required":[
                    "contract_id"
                ],
                "type":"object",
                "description":"Track live changes and trigger reanalysis automatically.",
                "output":{
                    "type":"null",
                    "description":"Starts real-time tracking of contract edits; no return value."
                }
            },
            {
                "name":"check_overall_compliance",
                "properties":{
                    "contract_id":{
                        "type":"string",
                        "description":"ID of the contract to validate for final approval."
                    }
                },

\end{minted}
\end{center}
\caption{RecordSynth: Sample Record - Part 4}
\label{lst:RecordSynthSampleRecord4}
\end{listing}

\begin{listing}[!htpb]
\begin{center}
\begin{minted}[
    gobble=4,
    fontsize=\scriptsize,
    breaklines=true,
]{json}

                "required":[
                    "contract_id"
                ],
                "type":"object",
                "description":"Validate if all flagged issues are resolved for final approval.",
                "output":{
                    "type":"boolean",
                    "description":"True if all compliance issues are resolved; false otherwise."
                }
            },
            {
                "name":"escalate_delayed_resolution",
                "properties":{
                    "contract_id":{
                        "type":"string",
                        "description":"Contract ID with unresolved high-risk clauses nearing deadlines."
                    }
                },
                "required":[
                    "contract_id"
                ],
                "type":"object",
                "description":"Automatically escalate unresolved high-risk clauses nearing deadlines.",
                "output":{
                    "type":"boolean",
                    "description":"True if escalation was successful for high-risk clauses nearing deadlines."
                }
            },
            {
                "name":"learn_from_resolution_outcomes",
                "properties":{
                    "review_outcome":{
                        "type":"string",
                        "description":"Outcome of the contract review (e.g., 'approved', 'rejected')."
                    }
                },
                "required":[
                    "review_outcome"
                ],
                "type":"object",
                "description":"Refine AI logic using feedback from final contract approvals.",
                "output":{
                    "type":"null",
                    "description":"Updates AI logic based on final contract approval outcomes; returns None."
                }
            }
        ],
        "pseudo_code":"function contract_compliance_agent(contract_id):\n    # Step 1: Begin real-time monitoring for live edits\n    monitor_real_time_edits(contract_id)\n\n    while monitoring_active:\n        # Step 2: Fetch latest contract version\n        current_draft = fetch_contract_draft(contract_id)\n        \n        # Step 3: Extract clauses and retrieve compliance criteria\n        clauses = analyze_contract_text(current_draft['text'])\n        jurisdiction = current_draft['jurisdiction']\n        laws = retrieve_jurisdictional_requirements(jurisdiction)\n        policies = get_internal_policies()\n        \n        flagged_issues = []\n        amendments = []\n        \n        # Step 4: Analyze each clause for compliance\n        for clause in clauses:\n            issues = check_clause_compliance(clause, laws, policies)\n            if issues:\n                flagged_issues.extend(issues)\n                \n                # Step 5: Generate improvement suggestions\n                for issue in issues:\n                    standardized_term = suggest_standardized_terms(issue.type, clause.type)\n                    amendments.append({\n                        'clause_id': clause.id,\n                        'original_text': clause.text,\n                        'suggestion': standardized_term\n                    })\n        \n        # Step 6: Record amendment proposals\n        for amendment in amendments:\n            log_contract_amendment(contract_id, amendment)\n        \n        # Step 7: Validate signatory authority\n        authority_valid = verify_signatory_authority(current_draft['signatory_role'], current_draft['scope'])\n        if not authority_valid:\n            flagged_issues.append(\"Signatory lacks required authority for this contract\")\n        \n        # Step 8: Generate compliance status report\n        compliance_summary = generate_compliance_summary(contract_id, flagged_issues)\n        \n        # Step 9: Alert legal team for unresolved critical issues\n        critical_issues = [issue for issue in flagged_issues if 'critical' in issue.severity]\n        if critical_issues:\n            alert_legal_team(contract_id, critical_issues)\n        \n        # Step 10: Final compliance validation\n        fully_compliant = check_overall_compliance(contract_id)\n        if not fully_compliant:\n            escalate_delayed_resolution(contract_id)  # Escalate near deadlines\n        else:\n            # Step 11: Post-approval learning\n            # Note: Actual outcome collection would occur after final approval\n            learn_from_resolution_outcomes('approved')  # Simplified placeholder",
        "possible_scenario":"AI escalates contract after failed authority check despite team inactivity",
        "user_utterance":"The procurement contract with TechSolutions has been pending for 4 days. Why hasn't it been approved yet?",

\end{minted}
\end{center}
\caption{RecordSynth: Sample Record - Part 5}
\label{lst:RecordSynthSampleRecord5}
\end{listing}

\begin{listing}[!htpb]
\begin{center}
\begin{minted}[
    gobble=4,
    fontsize=\scriptsize,
    breaklines=true,
]{json}

        "function_inputs_outputs":[
            {
                "function_call":"fetch_contract_draft('CT-9876')",
                "output":{
                    "id":"CT-9876",
                    "text":"Service Level Agreements... Payment terms... Confidentiality...",
                    "jurisdiction":"California",
                    "signatory_role":"Procurement Manager",
                    "scope":">$500k Vendor Agreements",
                    "last_edit":"2024-03-10T14:30:00Z"
                }
            },
            {
                "function_call":"analyze_contract_text(text)",
                "output":[
                    {
                        "id":"CLA-001",
                        "type":"Termination Clause",
                        "text":"...30-day notice..."
                    },
                    {
                        "id":"CLA-002",
                        "type":"Liability Cap",
                        "text":"$250k max liability"
                    }
                ]
            },
            {
                "function_call":"retrieve_jurisdictional_requirements('California')",
                "output":[
                    "Require written arbitration clauses",
                    "No liquidated damages limits exceeding statutory max"
                ]
            },
            {
                "function_call":"get_internal_policies()",
                "output":[
                    "All >$500k contracts require VP Legal sign-off",
                    "Standard confidentiality clauses v3.1"
                ]
            },
            {
                "function_call":"verify_signatory_authority('Procurement Manager', '>$500k Vendor Agreements')",
                "output":false
            },
            {
                "function_call":"generate_compliance_summary('CT-9876')",
                "output":{
                    "critical_issues":[
                        "Signatory lacks authority for $500k+ contracts (requires VP Legal)",
                        "Missing mandatory arbitration clause for CA jurisdiction"
                    ],
                    "suggestions":{
                        "CLA-002":"Suggest adding arbitration clause from policy v3.1"
                    }
                }
            },
            {
                "function_call":"check_overall_compliance('CT-9876')",
                "output":false
            },
            {
                "function_call":"escalate_delayed_resolution('CT-9876')",
                "output":true
            }
        ],
        "flow_summary":"The AI detected the $500k+ contract's signatory was only a Procurement Manager, violating policy requiring VP Legal approval. With no updates in 4 days and unresolved jurisdictional issues, the system escalated the contract to the Chief Legal Officer with full compliance failures outlined. Notifications were sent to legal stakeholders for urgent review.",
        "outcome":"The contract was automatically escalated to higher authority for approval after detecting both signatory authority failure and team inactivity, preventing non-compliance risks and ensuring critical contracts receive appropriate oversight despite delays.",
        
\end{minted}
\end{center}
\caption{RecordSynth: Sample Record - Part 6}
\label{lst:RecordSynthSampleRecord6}
\end{listing}

\begin{listing}[!htpb]
\begin{center}
\begin{minted}[
    gobble=4,
    fontsize=\scriptsize,
    breaklines=true,
]{json}
        
        "agentic_execution":[
            {
                "step_number":1,
                "functions_executed":[
                    "fetch_contract_draft('CT-9876')"
                ],
                "is_parallel":false,
                "description":"Fetch the latest version of the contract draft to initiate compliance checks",
                "input":[
                    "CT-9876"
                ],
                "output":[
                    {
                        "id":"CT-9876",
                        "text":"Service Level Agreements... Payment terms... Confidentiality...",
                        "jurisdiction":"California",
                        "signatory_role":"Procurement Manager",
                        "scope":">$500k Vendor Agreements",
                        "last_edit":"2024-03-10T14:30:00Z"
                    }
                ]
            },
            {
                "step_number":2,
                "functions_executed":[
                    "analyze_contract_text(text)"
                ],
                "is_parallel":false,
                "description":"Analyze the contract text to extract clauses for compliance review",
                "input":[
                    "Service Level Agreements... Payment terms... Confidentiality..."
                ],
                "output":[
                    [
                        {
                            "id":"CLA-001",
                            "type":"Termination Clause",
                            "text":"...30-day notice..."
                        },
                        {
                            "id":"CLA-002",
                            "type":"Liability Cap",
                            "text":"$250k max liability"
                        }
                    ]
                ]
            },
            {
                "step_number":3,
                "functions_executed":[
                    "retrieve_jurisdictional_requirements('California')",
                    "get_internal_policies()"
                ],
                "is_parallel":true,
                "description":"Retrieve jurisdictional legal requirements and internal policies in parallel",
                "input":[
                    [
                        "California"
                    ],
                    [

                    ]
                ],
                "output":[
                    [
                        "Require written arbitration clauses",
                        "No liquidated damages limits exceeding statutory max"
                    ],
                    [
                        "All >$500k contracts require VP Legal sign-off",
                        "Standard confidentiality clauses v3.1"
                    ]
                ]
            },

\end{minted}
\end{center}
\caption{RecordSynth: Sample Record - Part 7}
\label{lst:RecordSynthSampleRecord7}
\end{listing}

\begin{listing}[!htpb]
\begin{center}
\begin{minted}[
    gobble=4,
    fontsize=\scriptsize,
    breaklines=true,
]{json}
            
            {
                "step_number":4,
                "functions_executed":[
                    "verify_signatory_authority('Procurement Manager', '>$500k Vendor Agreements')"
                ],
                "is_parallel":false,
                "description":"Verify if the Procurement Manager has valid authority for contracts over $500k",
                "input":[
                    "Procurement Manager",
                    ">$500k Vendor Agreements"
                ],
                "output":[
                    false
                ]
            },
            {
                "step_number":5,
                "functions_executed":[
                    "generate_compliance_summary('CT-9876')"
                ],
                "is_parallel":false,
                "description":"Generate a compliance report highlighting critical issues like missing arbitration clause and signatory authority failure",
                "input":[
                    "CT-9876"
                ],
                "output":[
                    {
                        "critical_issues":[
                            "Signatory lacks authority for $500k+ contracts (requires VP Legal)",
                            "Missing mandatory arbitration clause for CA jurisdiction"
                        ],
                        "suggestions":{
                            "CLA-002":"Suggest adding arbitration clause from policy v3.1"
                        }
                    }
                ]
            },
            {
                "step_number":6,
                "functions_executed":[
                    "check_overall_compliance('CT-9876')"
                ],
                "is_parallel":false,
                "description":"Determine if the contract is fully compliant with all requirements",
                "input":[
                    "CT-9876"
                ],
                "output":[
                    false
                ]
            },
            {
                "step_number":7,
                "functions_executed":[
                    "escalate_delayed_resolution('CT-9876')"
                ],
                "is_parallel":false,
                "description":"Escalate the non-compliant contract to higher authority due to unresolved issues and inactivity",
                "input":[
                    "CT-9876"
                ],
                "output":[
                    true
                ]
            }
        ],
        "id":0
    },
]
\end{minted}
\end{center}
\caption{RecordSynth: Sample Record - Part 8}
\label{lst:RecordSynthSampleRecord8}
\end{listing}

\begin{listing}[!htpb]
\begin{center}
\begin{minted}[
    gobble=4,
    fontsize=\scriptsize,
    breaklines=true,
]{yaml}
    You are an expert AI assistant.
    
    Your task is to generate {{num}} unique and practical use cases for the given enterprise domain. The use cases should be relevant, realistic, and valuable in a business or technology context.
    
    Example:
    
    Domain : IT Service Management (ITSM)
    usecases : [
        "Automated Incident Routing: Use AI to categorize and route incoming incident tickets to the appropriate support team based on keywords and historical resolution data.",
        "Self-Service Portal with Knowledge Base: Enable users to resolve common IT issues (e.g., password reset, VPN access) through a self-service portal integrated with a dynamic knowledge base.",
        "Change Management Workflow Automation: Automate approval workflows for software updates and infrastructure changes, ensuring compliance and minimizing service disruption.",
        "IT Asset Lifecycle Tracking: Monitor and manage IT assets from procurement to disposal, including automatic updates to CMDB (Configuration Management Database).",
        "Proactive Problem Management: Identify recurring incidents using trend analysis and root cause detection to proactively eliminate systemic issues and reduce ticket volume."
    ]
    
    Now, generate {{num}} new and unique use cases for the following domain:
    
    {{domain}}
    
    Output Format:
        1. Only output a Python list (list[str]) containing the {{num}} use cases.
        2. Use plain English only. Do not use any other language, especially Chinese or Japanese.
        3. Ensure this output is different from any previous results.
        4. No preamble, explanation, or additional formatting—just the Python list.
\end{minted}
\end{center}
\caption{RecordSynth: Sample prompt for Creating Scenario}
\label{lst:pp1_create_scenario}
\end{listing}

\begin{listing}[!htpb]
\begin{center}
\begin{minted}[
    gobble=4,
    fontsize=\scriptsize,
    breaklines=true,
]{yaml}
    You are an expert AI assistant.

    Your task is to create a concise real time agentic scenario based on the domain and use case provided.

    Example:

    Domain : IT Service Management (ITSM)
    Use Case : Automated Incident Routing: Use AI to categorize and route incoming incident tickets to the appropriate support team based on keywords and historical resolution data.
    Agentic Scenario : A large enterprise IT helpdesk receives thousands of incident tickets daily through email, chat, and a self-service portal. To streamline response times, an AI-powered agent continuously monitors incoming tickets in real time. As each ticket arrives, the agent analyzes the subject line and description using natural language processing (NLP) to extract intent, urgency, and relevant keywords. It then references historical resolution data to determine the most likely resolution path and automatically routes the ticket to the appropriate support team—such as networking, application support, or end-user services—while tagging it with priority and suggested resolution steps. This reduces manual triage effort, accelerates issue resolution, and improves SLA compliance.

    Now, generate a new and unique use cases for the following domain:

    Domain : {{domain}}
    Use Case : {{use_case}}
    Agentic Scenario :

    Output Format:
        1. Output in a single paragraph
        2. Use plain English only. Do not use any other language, especially Chinese or Japanese.
        3. Ensure this output is different from any previous results.
        4. No preamble, explanation, or additional formatting.
\end{minted}
\end{center}
\caption{RecordSynth: Prompt for Creating a Real-Time Agentic Use Case}
\label{lst:pp1_create_real_time_usecase}
\end{listing}

\begin{listing}[!htpb]
\begin{center}
\begin{minted}[
    gobble=4,
    fontsize=\scriptsize,
    breaklines=true,
]{yaml}
    You are an expert AI assistant.

    Your task is to create a list of functions needed to implement the use case to cover all the possibilities.

    Example:

    Domain : IT Service Management (ITSM)
    Use Case : Automated Incident Routing: Use AI to categorize and route incoming incident tickets to the appropriate support team based on keywords and historical resolution data.
    Agentic Scenario : A large enterprise IT helpdesk receives thousands of incident tickets daily through email, chat, and a self-service portal. To streamline response times, an AI-powered agent continuously monitors incoming tickets in real time. As each ticket arrives, the agent analyzes the subject line and description using natural language processing (NLP) to extract intent, urgency, and relevant keywords. It then references historical resolution data to determine the most likely resolution path and automatically routes the ticket to the appropriate support team—such as networking, application support, or end-user services—while tagging it with priority and suggested resolution steps. This reduces manual triage effort, accelerates issue resolution, and improves SLA compliance.
    functions_list : ['fetch_pending_approval_requests() -> list # Retrieve incoming approval requests from the workflow queue.',
                    'evaluate_request_urgency(request) -> str # Analyze the request metadata to determine its urgency level.',
                    'get_organizational_policies(request_type) -> dict # Retrieve routing rules and escalation policies for a specific request type.',
                    'get_approver_candidates(request) -> list # List potential approvers based on role, department, and permissions.',
                    'fetch_approver_workload(approver_id) -> dict # Get current workload (open tasks, approvals, meetings) of an approver.',
                    'score_approvers(candidates, workload_data, urgency, policy_rules) -> list # Score and rank approvers based on policy alignment, availability, and urgency.',
                    'select_best_approver(scored_candidates) -> str # Select the top-ranked approver based on scores and routing logic.',
                    'assign_request_to_approver(request_id, approver_id) -> bool # Assign the approval request to the selected approver.',
                    'notify_approver(approver_id, request_id) -> None # Send real-time notification to the approver about the assigned request.',
                    'log_routing_decision(request_id, approver_id, reasoning) -> None # Log the decision trail for audit and transparency.',
                    'escalate_if_no_approval_in_time(request_id) -> bool # Monitor for SLA breaches and escalate requests proactively.',
                    'learn_from_feedback(routing_outcome, approver_feedback) -> None # Update the model or logic based on outcome quality and user feedback.']

    Now, generate a new functions_list for the following:

    Domain : {{domain}}
    Use Case : {{use_case}}
    Agentic Scenario : {{agentic_scenario}}
    functions_list :

    Instructions:
        1. Output must be a Python list of functions like: 'function_name(parameter1, ...) -> return_type # Description of what the function does.'
        2. Each function must include a return type using simple types only: str, bool, list, dict, or None. Avoid complex or custom types.
        3. The list must be named functions_list = [...].
        4. Use plain English only. Do not use any other language, especially Chinese or Japanese.
        5. Ensure this output is different from any previous results.
        6. No preamble, explanation, or additional formatting.
\end{minted}
\end{center}
\caption{RecordSynth: Prompt for Creating the Functions List}
\label{lst:pp1_create_functions_list}
\end{listing}

\begin{listing}[!htpb]
\begin{center}
\begin{minted}[
    gobble=4,
    fontsize=\scriptsize,
    breaklines=true,
]{yaml}
    You are an expert AI assistant.

    Your task is to create a pseudo code using the given functions by covering all the possibilities in the use case.

    Example:

    Domain : IT Service Management (ITSM)
    Use Case : Automated Incident Routing: Use AI to categorize and route incoming incident tickets to the appropriate support team based on keywords and historical resolution data.
    Agentic Scenario : A large enterprise IT helpdesk receives thousands of incident tickets daily through email, chat, and a self-service portal. To streamline response times, an AI-powered agent continuously monitors incoming tickets in real time. As each ticket arrives, the agent analyzes the subject line and description using natural language processing (NLP) to extract intent, urgency, and relevant keywords. It then references historical resolution data to determine the most likely resolution path and automatically routes the ticket to the appropriate support team—such as networking, application support, or end-user services—while tagging it with priority and suggested resolution steps. This reduces manual triage effort, accelerates issue resolution, and improves SLA compliance.
    ffunctions_list : ['fetch_pending_approval_requests() -> list # Retrieve incoming approval requests from the workflow queue.',
                    'evaluate_request_urgency(request) -> str # Analyze the request metadata to determine its urgency level.',
                    'get_organizational_policies(request_type) -> dict # Retrieve routing rules and escalation policies for a specific request type.',
                    'get_approver_candidates(request) -> list # List potential approvers based on role, department, and permissions.',
                    'fetch_approver_workload(approver_id) -> dict # Get current workload (open tasks, approvals, meetings) of an approver.',
                    'score_approvers(candidates, workload_data, urgency, policy_rules) -> list # Score and rank approvers based on policy alignment, availability, and urgency.',
                    'select_best_approver(scored_candidates) -> str # Select the top-ranked approver based on scores and routing logic.',
                    'assign_request_to_approver(request_id, approver_id) -> bool # Assign the approval request to the selected approver.',
                    'notify_approver(approver_id, request_id) -> None # Send real-time notification to the approver about the assigned request.',
                    'log_routing_decision(request_id, approver_id, reasoning) -> None # Log the decision trail for audit and transparency.',
                    'escalate_if_no_approval_in_time(request_id) -> bool # Monitor for SLA breaches and escalate requests proactively.',
                    'learn_from_feedback(routing_outcome, approver_feedback) -> None # Update the model or logic based on outcome quality and user feedback.']

    pseudo_code :
                function dynamic_approval_routing_agent():
                    # Step 1: Get all pending approval requests
                    approval_requests = fetch_pending_approval_requests()

                    for request in approval_requests:
                        # Step 2: Analyze the urgency of the request
                        urgency = evaluate_request_urgency(request)

                        # Step 3: Get applicable policy rules
                        policy_rules = get_organizational_policies(request.type)

                        # Step 4: Fetch eligible approvers based on request context
                        approver_candidates = get_approver_candidates(request)

                        # If no candidates, raise flag and log
                        if approver_candidates is empty:
                            log_routing_decision(request.id, None, "No eligible approvers found.")
                            continue

                        # Step 5: Gather current workload for each candidate
                        workload_data = {}
                        for approver in approver_candidates:
                            workload_data[approver.id] = fetch_approver_workload(approver.id)

                        # Step 6: Score candidates using workload, urgency, and policy
                        scored_candidates = score_approvers(approver_candidates, workload_data, urgency, policy_rules)

\end{minted}
\end{center}
\caption{RecordSynth: Prompt for Creating Pseudo Code - Part 1}
\label{lst:pp1_create_pseudo_code_pt1}
\end{listing}

\begin{listing}[!htpb]
\begin{center}
\begin{minted}[
    gobble=4,
    fontsize=\scriptsize,
    breaklines=true,
]{yaml}

                        # Step 7: Pick the best approver
                        selected_approver = select_best_approver(scored_candidates)

                        # Fallback if no approver meets threshold
                        if selected_approver is None:
                            # Escalate or fallback based on policy
                            fallback_approver = policy_rules.get("fallback_approver")
                            if fallback_approver:
                                selected_approver = fallback_approver
                                reason = "Fallback approver selected due to no qualified candidates."
                            else:
                                log_routing_decision(request.id, None, "No suitable approver available; manual intervention required.")
                                continue
                        else:
                            reason = "Selected based on dynamic scoring."

                        # Step 8: Assign request and notify approver
                        assign_request_to_approver(request.id, selected_approver.id)
                        notify_approver(selected_approver.id, request.id)

                        # Step 9: Log the routing decision
                        log_routing_decision(request.id, selected_approver.id, reason)

                        # Step 10: Monitor SLA and escalate if needed
                        monitor_sla_and_escalate(request.id)

                        # Step 11: Learn from outcome/feedback after resolution
                        routing_outcome = wait_for_approval_outcome(request.id)
                        feedback = collect_approver_feedback(request.id)
                        learn_from_feedback(routing_outcome, feedback)


    Now, generate a new functions_list for the following:

    Domain : {{domain}}
    Use Case : {{use_case}}
    Agentic Scenario : {{agentic_scenario}}
    functions_list : {{function_list}}
    pseudo_code :

    Output Format:
        1. Output must be a pydantic code that implements the agentic scenario. 
        2. Use plain English only. Do not use any other language, especially Chinese or Japanese.
        3. Ensure this output is different from any previous results.
        4. No preamble, explanation, or additional formatting.
        5. Make sure it must be a valid python code.
        6. Make sure it should use only the functions and arguments provided in the functions_list.
\end{minted}
\end{center}
\caption{RecordSynth: Prompt for Creating Pseudo Code - Part 2}
\label{lst:pp1_create_pseudo_code_pt2}
\end{listing}

\begin{listing}[!htpb]
\begin{center}
\begin{minted}[
    gobble=4,
    fontsize=\scriptsize,
    breaklines=true,
]{yaml}
    You are an expert AI assistant.

    Your task is to create a list of {{num}} possible scenarios by considering the agentic scenario, function_list and pseudo code provided.

    Example:

    Agentic Scenario : A large enterprise IT helpdesk receives thousands of incident tickets daily through email, chat, and a self-service portal. To streamline response times, an AI-powered agent continuously monitors incoming tickets in real time. As each ticket arrives, the agent analyzes the subject line and description using natural language processing (NLP) to extract intent, urgency, and relevant keywords. It then references historical resolution data to determine the most likely resolution path and automatically routes the ticket to the appropriate support team—such as networking, application support, or end-user services—while tagging it with priority and suggested resolution steps. This reduces manual triage effort, accelerates issue resolution, and improves SLA compliance.

    functions_list : ['fetch_pending_approval_requests()  # Retrieve incoming approval requests from the workflow queue.',
        'evaluate_request_urgency(request)  # Analyze the request metadata to determine its urgency level.',
        'get_organizational_policies(request_type)  # Retrieve routing rules and escalation policies for a specific request type.',
        'get_approver_candidates(request)  # List potential approvers based on role, department, and permissions.',
        'fetch_approver_workload(approver_id)  # Get current workload (open tasks, approvals, meetings) of an approver.',
        'score_approvers(candidates, workload_data, urgency, policy_rules)  # Score and rank approvers based on policy alignment, availability, and urgency.',
        'select_best_approver(scored_candidates)  # Select the top-ranked approver based on scores and routing logic.',
        'assign_request_to_approver(request_id, approver_id)  # Assign the approval request to the selected approver.',
        'notify_approver(approver_id, request_id)  # Send real-time notification to the approver about the assigned request.',
        'log_routing_decision(request_id, approver_id, reasoning)  # Log the decision trail for audit and transparency.',
        'escalate_if_no_approval_in_time(request_id)  # Monitor for SLA breaches and escalate requests proactively.',
        'learn_from_feedback(routing_outcome, approver_feedback)  # Update the model or logic based on outcome quality and user feedback.']
        
    pseudo_code : 
                function dynamic_approval_routing_agent():
                    # Step 1: Get all pending approval requests
                    approval_requests = fetch_pending_approval_requests()

                    for request in approval_requests:
                        # Step 2: Analyze the urgency of the request
                        urgency = evaluate_request_urgency(request)

                        # Step 3: Get applicable policy rules
                        policy_rules = get_organizational_policies(request.type)

                        # Step 4: Fetch eligible approvers based on request context
                        approver_candidates = get_approver_candidates(request)

                        # If no candidates, raise flag and log
                        if approver_candidates is empty:
                            log_routing_decision(request.id, None, "No eligible approvers found.")
                            continue

                        # Step 5: Gather current workload for each candidate
                        workload_data = {}
                        for approver in approver_candidates:
                            workload_data[approver.id] = fetch_approver_workload(approver.id)

                        # Step 6: Score candidates using workload, urgency, and policy
                        scored_candidates = score_approvers(approver_candidates, workload_data, urgency, policy_rules)

                        # Step 7: Pick the best approver
                        selected_approver = select_best_approver(scored_candidates)

                        # Fallback if no approver meets threshold
                        if selected_approver is None:
                            # Escalate or fallback based on policy
                            fallback_approver = policy_rules.get("fallback_approver")
                            if fallback_approver:
                                selected_approver = fallback_approver
                                reason = "Fallback approver selected due to no qualified candidates."
                            else:
                                log_routing_decision(request.id, None, "No suitable approver available; manual intervention required.")
                                continue
                        else:
                            reason = "Selected based on dynamic scoring."

\end{minted}
\end{center}
\caption{RecordSynth: Prompt for Creating Comprehensive Scenarios - Part 1}
\label{lst:pp1_create_comprehensive_scenarios_pt1}
\end{listing}

\begin{listing}[!htpb]
\begin{center}
\begin{minted}[
    gobble=4,
    fontsize=\scriptsize,
    breaklines=true,
]{yaml}

                        # Step 8: Assign request and notify approver
                        assign_request_to_approver(request.id, selected_approver.id)
                        notify_approver(selected_approver.id, request.id)

                        # Step 9: Log the routing decision
                        log_routing_decision(request.id, selected_approver.id, reason)

                        # Step 10: Monitor SLA and escalate if needed
                        monitor_sla_and_escalate(request.id)

                        # Step 11: Learn from outcome/feedback after resolution
                        routing_outcome = wait_for_approval_outcome(request.id)
                        feedback = collect_approver_feedback(request.id)
                        learn_from_feedback(routing_outcome, feedback)

    possible_scenarios : [
        "Request routed to best-fit approver based on workload and urgency",
        "No eligible approvers found for request type",
        "Top candidate is overloaded, fallback approver assigned",
        "Multiple candidates available, selected based on lowest workload",
        "Urgent request routed to next available qualified approver despite policy preference",
        "Request assigned successfully, approver notified in real-time",
        "Approver does not respond within SLA, request escalated",
        "Approver declines or fails to act, rerouting triggered automatically",
        "Approval request auto-assigned to backup approver due to policy rule",
        "Approval routing fails due to missing or invalid policy configuration",
        "Request type is unsupported, logged for manual intervention",
        "Agent detects loop or repeated failure in routing, triggers admin alert",
        "Feedback collected from approver on inappropriate assignment",
        "Routing agent updates decision logic based on post-approval feedback",
        "Request marked high-priority, bypasses normal routing path",
        "Scoring logic adjusted due to real-time change in workload data",
        "Approval assigned but rejected, agent triggers re-routing or clarification request",
        "Agent simulates scoring before final assignment for audit traceability",
        "Approver pool updated dynamically due to org structure change",
        "Routing delayed due to dependency (e.g., pending prerequisite approval)",
        "Multiple requests batched and routed to optimize approver load",
        "Agent auto-generates explanation for routing decision for audit logs",
        "System outage occurs mid-routing, agent retries or queues request",
        "Approver out-of-office, auto-redirect to delegate",
        "Urgency misclassified, agent receives corrective feedback",
        "Agent learns from escalated approvals to better prioritize future ones",
        "Load spikes trigger adaptive routing to non-preferred approvers",
        "Approval is rerouted due to SLA breach before initial response",
        "Agent overrides human-assigned approver due to updated rules",
        "User manually reassigns request; agent logs override and feedback"
    ]

    Now, generate a list of {{num}} possible scenarios by considering the following:

    Agentic Scenario : {{agentic_scenario}}
    functions_list : {{function_list}}
    pseudo_code : {{pseudo_code}}
    possible_scenarios :

    Output Format:
        1. Only output a Python list (list[str]) containing the {{num}} possible scenarios
        2. Use plain English only. Do not use any other language, especially Chinese or Japanese.
        3. Ensure this output is different from any previous results.
        4. No preamble, explanation, or additional formatting—just the Python list.
\end{minted}
\end{center}
\caption{RecordSynth: Prompt for Creating Comprehensive Scenarios - Part 2}
\label{lst:pp1_create_comprehensive_scenarios_pt2}
\end{listing}

\begin{listing}[!htpb]
\begin{center}
\begin{minted}[
    gobble=4,
    fontsize=\scriptsize,
    breaklines=true,
]{yaml}
        You are an expert AI assistant.
    
        Given the scenario, policy and list of function_list, below is your task.
    
            1. Create a user utterance. 
            2. Based on the user utterance and policy and available function_list, create function inputs.
            3. Based on the function inputs / outputs and policy, generate function expected outputs
            4. Also based on the all above, create a flow summary & outcome.
    
        Example:
    
        scenario : Request routed to best-fit approver based on workload and urgency
    
        policy : An AI agent monitors incoming approval requests across departments and dynamically routes them to the most appropriate approver in real time, based on organizational  policy rules, the current workload of potential approvers, and the urgency level of each request. For example, if a high-priority purchase order needs rapid approval and the usual approver is overloaded, the agent autonomously reassigns it to a qualified backup approver with available capacity, ensuring compliance and speed without human intervention.
    
        functions_list : ['fetch_pending_approval_requests() -> list # Retrieve incoming approval requests from the workflow queue.',
                        'evaluate_request_urgency(request) -> str # Analyze the request metadata to determine its urgency level.',
                        'get_organizational_policies(request_type) -> dict # Retrieve routing rules and escalation policies for a specific request type.',
                        'get_approver_candidates(request) -> list # List potential approvers based on role, department, and permissions.',
                        'fetch_approver_workload(approver_id) -> dict # Get current workload (open tasks, approvals, meetings) of an approver.',
                        'score_approvers(candidates, workload_data, urgency, policy_rules) -> list # Score and rank approvers based on policy alignment, availability, and urgency.',
                        'select_best_approver(scored_candidates) -> str # Select the top-ranked approver based on scores and routing logic.',
                        'assign_request_to_approver(request_id, approver_id) -> bool # Assign the approval request to the selected approver.',
                        'notify_approver(approver_id, request_id) -> None # Send real-time notification to the approver about the assigned request.',
                        'log_routing_decision(request_id, approver_id, reasoning) -> None # Log the decision trail for audit and transparency.',
                        'escalate_if_no_approval_in_time(request_id) -> bool # Monitor for SLA breaches and escalate requests proactively.',
                        'learn_from_feedback(routing_outcome, approver_feedback) -> None # Update the model or logic based on outcome quality and user feedback.']
            
        pseudo_code : 
                    function dynamic_approval_routing_agent():
                        # Step 1: Get all pending approval requests
                        approval_requests = fetch_pending_approval_requests()
    
                        for request in approval_requests:
                            # Step 2: Analyze the urgency of the request
                            urgency = evaluate_request_urgency(request)
    
                            # Step 3: Get applicable policy rules
                            policy_rules = get_organizational_policies(request.type)
    
                            # Step 4: Fetch eligible approvers based on request context
                            approver_candidates = get_approver_candidates(request)
    
                            # If no candidates, raise flag and log
                            if approver_candidates is empty:
                                log_routing_decision(request.id, None, "No eligible approvers found.")
                                continue
    
                            # Step 5: Gather current workload for each candidate
                            workload_data = {}
                            for approver in approver_candidates:
                                workload_data[approver.id] = fetch_approver_workload(approver.id)
    
                            # Step 6: Score candidates using workload, urgency, and policy
                            scored_candidates = score_approvers(approver_candidates, workload_data, urgency, policy_rules)
\end{minted}
\end{center}
\caption{RecordSynth: Prompt for Creating the User Scenario Pack - Part 1}
\label{lst:pp1_create_user_scenario_pt1}
\end{listing}

\begin{listing}[!htpb]
\begin{center}
\begin{minted}[
    gobble=4,
    fontsize=\scriptsize,
    breaklines=true,
]{yaml}

                            # Step 7: Pick the best approver
                            selected_approver = select_best_approver(scored_candidates)
    
                            # Fallback if no approver meets threshold
                            if selected_approver is None:
                                # Escalate or fallback based on policy
                                fallback_approver = policy_rules.get("fallback_approver")
                                if fallback_approver:
                                    selected_approver = fallback_approver
                                    reason = "Fallback approver selected due to no qualified candidates."
                                else:
                                    log_routing_decision(request.id, None, "No suitable approver available; manual intervention required.")
                                    continue
                                    
                            else:
                                reason = "Selected based on dynamic scoring."
    
                            # Step 8: Assign request and notify approver
                            assign_request_to_approver(request.id, selected_approver.id)
                            notify_approver(selected_approver.id, request.id)
    
                            # Step 9: Log the routing decision
                            log_routing_decision(request.id, selected_approver.id, reason)
    
                            # Step 10: Monitor SLA and escalate if needed
                            monitor_sla_and_escalate(request.id)
    
                            # Step 11: Learn from outcome/feedback after resolution
                            routing_outcome = wait_for_approval_outcome(request.id)
                            feedback = collect_approver_feedback(request.id)
                            learn_from_feedback(routing_outcome, feedback)
    
    
        utterance : Hey, I just submitted a high-priority purchase request for $50,000 for the marketing team. Can you make sure it gets approved quickly?
    
        function_inputs_outputs : fetch_pending_approval_requests() → Returns a list of dicts (list[dict]):
                            [
                                {
                                    "id": "PR-5478",                        # str
                                    "type": "purchase_request",             # str
                                    "department": "Marketing",              # str
                                    "amount": 50000,                        # int
                                    "submitted_by": "user123",              # str
                                    "submitted_at": "2025-07-07T10:15:00Z"  # str (ISO 8601 datetime)
                                }
                            ]
    
                            evaluate_request_urgency(request) → Evaluates based on amount, department, and business rules
                            → Output: urgency (str)  
                            Example: urgency = "High"
    
                            get_organizational_policies("purchase_request") → Returns a policy config dict (dict):
                            {
                                "max_amount_thresholds": {                 # dict[str, int]
                                    "manager": 10000,
                                    "director": 25000,
                                    "vp": 75000
                                },
                                "fallback_approver": "vp_approver_02",     # str
                                "workload_threshold": 80,                  # int (%)
                                "prefer_same_department": True             # bool
                            }
    
                            get_approver_candidates(request) → Returns a list of approver dicts (list[dict]):
                            [
                                {"id": "manager_01", "role": "manager", "dept": "Marketing"},   # all str
                                {"id": "director_01", "role": "director", "dept": "Marketing"},
                                {"id": "vp_01", "role": "vp", "dept": "Finance"}
                            ]

\end{minted}
\end{center}
\caption{RecordSynth: Prompt for Creating the User Scenario Pack - Part 2}
\label{lst:pp1_create_user_scenario_pt2}
\end{listing}

\begin{listing}[!htpb]
\begin{center}
\begin{minted}[
    gobble=4,
    fontsize=\scriptsize,
    breaklines=true,
]{yaml}

                            fetch_approver_workload(approver_id) → Returns a workload map dict (dict[str, int]):
                            {
                            "manager_01": 90, # too busy # int (percentage workload)
                            "director_01": 45, # acceptable
                            "vp_01": 75 # borderline
                            }
    
                            score_approvers(candidates, workload_data, urgency, policy_rules) → Returns a ranked list (list[dict]):
                            [
                                {"id": "director_01", "score": 0.95},  # id: str, score: float
                                {"id": "vp_01", "score": 0.65}
                            ]
    
        flow_summary : The AI agent detects the incoming high-priority purchase request. It identifies that the request amount exceeds the manager’s limit and evaluates urgency as “High”. Based on policy, it looks for director or VP-level approvers. The director in the same department has low workload (45%), making them the best-fit candidate. The agent assigns the request to director_01, notifies them in real-time, and logs the routing decision with justification. SLA monitoring starts in the background to ensure timely action.
    
        outcome : The approval request is dynamically routed to the most capable and available person, balancing compliance, efficiency, and business urgency — all without human intervention.
    
        Now, generate your response for the below:
    
        scenario : {{scenario}}
        policy : {{policy}}
        function_list : {{function_list}}
        pseudo_code : {{pseudo_code}}
        utterance : 
        function_inputs_outputs : 
        flow_summary : 
        outcome : 
    
    
        Output Format:
            1. Only output in a valid json format with the schema like {"utterance": "", "function_inputs_outputs": [], "flow_summary": "", "outcome": ""}
            2. Use plain English only. Do not use any other language, especially Chinese or Japanese.
            3. Ensure all the keys are values should be present as mentioned in the schema.
            4. function_inputs_outputs should be a list of dictionaries.
            5. No preamble, explanation, or additional formatting—just the valid json and it should be loaded in json.loads() without any failure.
            6. Make sure the function_inputs_outputs list contains the correct function calls and their outputs based on the scenario, pseudo_code and function_list.
\end{minted}
\end{center}
\caption{RecordSynth: Prompt for Creating the User Scenario Pack - Part 3}
\label{lst:pp1_create_user_scenario_pt3}
\end{listing}

\begin{listing}[!htpb]
\begin{center}
\begin{minted}[
    gobble=4,
    fontsize=\scriptsize,
    breaklines=true,
]{yaml}
        You are an expert AI assistant.
        You will receive a list of functions, its input and output parameters and pseudo code.
        convert the given function list to a json format given below.
    
        Example : 
        input_function : [detect_network_congestion(packet_data, threshold_latency, threshold_packet_loss) -> bool # Analyzes network traffic data to detect signs of congestion based on latency and packet loss thresholds.]
        expected_function : [{
                                "name": "detect_network_congestion",
                                "properties": {
                                    "packet_data": {
                                        "description": "List of dictionaries containing network packet information.",
                                        "items": {
                                            "additionalProperties": true,
                                            "type": "object"
                                        },
                                        "type": "array"
                                    },
                                    "threshold_latency": {
                                        "default": 100.0,
                                        "description": "Threshold latency in milliseconds.",
                                        "minimum": 0,
                                        "type": "number"
                                    },
                                    "threshold_packet_loss": {
                                        "default": 5.0,
                                        "description": "Threshold packet loss percentage.",
                                        "minimum": 0,
                                        "type": "number"
                                    }
                                },
                                "required": [
                                    "packet_data"
                                ],
                                "type": "object",
                                "description": "Analyzes network traffic data to detect signs of congestion based on latency and packet loss thresholds."
                            },"output": {
                                "type": "boolean",
                                "description": "Returns true if the network is congested based on the thresholds, otherwise false."
                            }]
    
        now, generate your response for the below:
    
        function_list : {{function_list}}
        pseudo_code : {{pseudo_code}}
        function_inputs_outputs : {{function_inputs_outputs}}
        expected_function : 
    
        Output Format:
            1. Only output in a valid json format with the schema as list of <key> : <value>
            2. Use plain English only. Do not use any other language, especially Chinese or Japanese.
            3. Ensure all the keys are values should be present as mentioned in the schema.
            4. No preamble, explanation, or additional formatting—just the Python list.
\end{minted}
\end{center}
\caption{RecordSynth: Prompt for Creating Function Descriptions (JSON Specifications)}
\label{lst:pp1_create_function_descriptions}
\end{listing}

\begin{listing}[!htpb]
\begin{center}
\begin{minted}[
    gobble=4,
    fontsize=\scriptsize,
    breaklines=true,
]{yaml}
        You are an expert AI assistant.
        Given the utterance, policy, function_inputs_outputs, flow_summary  and outcome, can you come up with a steps of the functions to be executed to get the desired outcome? 
        Make sure some functions can be called in parallel and some functions can be called in sequence based on the policy. 
        If functions can be called in parallel, put them in a single step.
        If the step contains parallel functions (more than 1 functions per step), make sure is_parallel is true. 
        If the step contains sequential (only one function per step), make sure is_parallel is false.
    
        Utterance: {{utterance}}
        Policy: {{policy}}
        Function_inputs_outputs: {{function_inputs_outputs}}
        Flow summary: {{flow_summary}}
        Outcome: {{outcome}}
    
        Steps : [{"step_number": 1, "functions_executed": ["function1", "function2"], "is_parallel": true/false, 
                "description": "Description of the step including the function call and the input parameters and the output",
                "input": ["Input parameters for the function call",],
                "output": ["Output of the function call"}, ...]
    
        Output Format:
            1. Only output in a valid json format with the schema as list of {Step_number, functions_executed, is_parallel, description, input, output}
            2. Ensure the input and output are list where each element corresponds to the functions executed at that step.
            2. Use plain English only. Do not use any other language, especially Chinese or Japanese.
            3. Ensure all the keys are values should be present as mentioned in the schema.
            4. No preamble, explanation, or additional formatting—just the valid json and it should be loaded in json.loads() without any failure.
            5. Make sure the steps (order of functions execution) should be as per the pseudo_code, flow summary and the outcome. Think logical and come up with the right order.
            6. Ensure the functions called and their input and output in the each step are strictly as per the pseudo code and Function_inputs_outputs.
            7. Ensure the parallel and sequential execution is as per the policy and parallel executions are grouped together in a step.
            
\end{minted}
\end{center}
\caption{RecordSynth: Prompt for Creating Agentic Execution Steps}
\label{lst:pp1_create_agentic_execution}
\end{listing}

\begin{listing}[!htpb]
\begin{center}
\begin{minted}[
    gobble=4,
    fontsize=\scriptsize,
    breaklines=true,
]{yaml}
    You are an expert workflow generator for tool-based DAGs.

    **Given:**
    - DAG Type: {{dag_type}}
    - Description: {{dag_type_description}}
    - Structure Template: {{dag_type_structure}}
    - Available Tools (with input/output fields): {{function_sequence}}

    **Task:**
    Generate a JSON array representing a workflow DAG that:
    - Uses the provided tools.
    - Follows the structure of the DAG Type template (replace T1, T2, etc. with actual tool names).
    - Wires each tool’s inputs according to the outputs of previous tools as per the structure.
    - Includes the "from" field: use "user" if the tool gets input from the user; otherwise, specify the previous tool name(s) it depends on.
    - For each tool, specify its required "inputs", mapping them appropriately to previous outputs 
      (e.g., "fetch_weather_forecast.forecast" for a tool needing weather data).
    - Always output a VALID JSON (list/array of dictionaries), even if there is only one step/tool in the workflow.
    - Do not output a single dictionary by itself—always wrap it in a list.

    **Example Output Format:**

    Below are several correct examples for different DAG types:

    Linear chain:
    {{dag_example_1}}

    Diamond:
    {{dag_example_2}}

    Fan-in:
    {{dag_example_3}}

    Single-step (must be a list with one dictionary):
    {{dag_example_4}}

    Output Format:
        1. Output must be valid JSON that loads via json.loads().
        2. Use plain English only. Do not use any other language.
        3. Wrap the entire DAG in a list of dictionaries.
        4. No preamble, explanation, or extra formatting—just the valid JSON list.
\end{minted}
\end{center}
\caption{DAGFirstGeneration: Prompt for Generating DAG Structure from Template}
\label{lst:pp3_generate_dag_from_template}
\end{listing}

\begin{listing}[!htpb]
\begin{center}
\begin{minted}[
    gobble=4,
    fontsize=\scriptsize,
    breaklines=true,
]{yaml}
    Generate a single, natural, and highly specific user utterance that would **uniquely** trigger execution of the provided workflow DAG.

    The utterance must include or imply all information needed to fill every input for every step in the DAG,
    including values for every key in any dictionary input. It must be so specific and comprehensive that,
    given the available tools and the information in the utterance, **no other possible DAG or workflow could resolve the request—only this exact DAG can answer it completely.**

    If a DAG step requires a dictionary input with specific keys, the utterance must naturally provide a value for every required key,
    so the workflow can be executed without asking the user for any further details.

    The result should be a single, conversational question or request—not a list, form, or template.

    ## INSTRUCTIONS:
    Given the following workflow DAG, write a natural-sounding, highly specific user utterance that would trigger its execution.
    - The utterance must cover all required input fields for all steps in the DAG, with enough detail to construct any dictionaries or values needed.
    - If the input is a dictionary with keys, include (in the utterance) information for every key.
    - **Make the utterance so uniquely detailed that, given the tools and info, only this DAG can answer the question completely and correctly.**
    - Do not output anything except a single natural language utterance.

    ## EXAMPLES:

    ### Example 1: Linear chain
    DAG:
    {{dag_example_1}}
    User Utterance:
      "For the next 5 days in Austin, taking into account the detailed weather forecast, send me a clothing recommendation and notify me with the suggested outfit each day."

    ### Example 2: Diamond
    DAG:
    {{dag_example_2}}
    User Utterance:
      "Retrieve the article from https://example.com/news_article.pdf, summarize its content, analyze its sentiment, and then compile both the summary and sentiment analysis into a single report for me."

    ### Example 3: Fan-in
    DAG:
    {{dag_example_3}}
    User Utterance:
      "Using both the latest price of Apple stock and today’s news headlines about Apple, give me a comprehensive investment recommendation for AAPL."

    ### Example 4: Single-step
    DAG:
    {{dag_example_4}}
    User Utterance:
      "What is the current local time in Paris, France, according to the Europe/Paris time zone?"

    ---
    Now, given the following DAG, generate a single, natural, and uniquely specific user utterance that contains or implies all the required information for every input field and dictionary key—such that only this workflow DAG could fulfill the request:

    DAG:
    {{dag}}
\end{minted}
\end{center}
\caption{DAGFirstGeneration: Prompt for Generating User Utterance from DAG}
\label{lst:pp3_llm_dag_user_utterance_generator}
\end{listing}

\begin{listing}[!htpb]
\begin{center}
\begin{minted}[
    gobble=4,
    fontsize=\scriptsize,
    breaklines=true,
]{yaml}
    You are a validation assistant responsible for verifying whether a given tool-based DAG can be executed based on a user utterance.

    Below is the user utterance from the user:
    {{user_utterance}}

    And here is the DAG that the AI agent plans to use to answer this instruction:
    {{dag}}

    Your task is to evaluate whether this DAG is **executable**, based only on whether the user has provided all necessary user-supplied inputs.

    **Additional requirement:**  
    The user utterance must be so uniquely specific and detailed, given the available tools and information, that only this exact DAG (and no other workflow) could fully and correctly answer the user’s request.

    Guidelines:
    - Go step by step through the DAG.
    - For each tool in the DAG, identify which inputs are:
        - a) provided by previous tools (intermediate), or
        - b) required directly from the user.
    - If any input required by **any tool** in the DAG (even tools at later stages) is **not available from an earlier tool**, it **must be supplied by the user** in their utterance.
    - If even one such input is missing or ambiguous, mark the DAG as **not executable**.
    - In addition, consider whether the user utterance is so detailed and tailored that, given the tools and info, **no other DAG could fully satisfy the request**. If a less specific DAG could also answer, mark as not executable and explain.

    Respond strictly in the following JSON format:
    {
      "DAG_Executable": true or false,
      "DAG_Executable_Description": "Explain briefly why it is executable or not, referring to the specific inputs that are missing, ambiguous, or if the utterance could be resolved by another DAG."
    }

    Instructions:
    - Do NOT include markdown syntax (e.g., no ```json).
    - Do NOT include any explanatory text before or after the JSON block.
    - Be concise and concrete in the description.
\end{minted}
\end{center}
\caption{DAGFirstGeneration: Prompt for DAG Executability Validation from User Utterance}
\label{lst:pp3_dag_judge}
\end{listing}

\begin{listing}[!htpb]
\begin{center}
\begin{minted}[
    gobble=4,
    fontsize=\scriptsize,
    breaklines=true,
]{yaml}
    You are a helpful AI assistant tasked with correcting an incomplete or insufficient user utterance so that an AI agent can fully and uniquely execute its workflow DAG.

    The agent attempted to execute the workflow but failed due to missing or ambiguous user-supplied inputs.

    === ORIGINAL USER UTTERANCE ===
    {{user_utterance}}

    === EXECUTION ERROR ===
    {{DAG_Executable_Description}}

    === DAG ===
    {{dag}}

    Your task:
    - Generate a corrected version of the user utterance that:
      - **Fixes the issue described above (for example, adds missing details or clarifies ambiguous ones as required by the error explanation)**
      - Preserves the original user's intent and tone
      - **Includes or implies all required information needed to fill every input for every step in the DAG, including values for every key in any dictionary input**
      - Makes the utterance so specific and comprehensive that, given the available tools and the information in the utterance, **no other possible DAG or workflow could resolve the request—only this exact DAG can answer it completely**
      - Presents all details in a natural, conversational way (not a list or form)
      - Does not mention tools, DAGs, or technical terms—only sound like a user making a natural request

    ## INSTRUCTIONS:
    - Ensure the utterance covers all required input fields for all steps in the DAG, with enough detail to construct any dictionaries or values needed.
    - If the input is a dictionary with keys, include (in the utterance) information for every key.
    - Make the utterance so uniquely detailed that, given the tools and info, only this DAG can answer the question completely and correctly.
    - Only output a single natural language utterance (no lists, no bullet points, no commentary).

    ## EXAMPLES:

    ### Example 1: Linear chain
    DAG:
    {{dag_example_1}}
    User Utterance:
      "For the next 5 days in Austin, taking into account the detailed weather forecast, send me a clothing recommendation and notify me with the suggested outfit each day."

    ### Example 2: Diamond
    DAG:
    {{dag_example_2}}
    User Utterance:
      "Retrieve the article from https://example.com/news_article.pdf, summarize its content, analyze its sentiment, and then compile both the summary and sentiment analysis into a single report for me."

    ### Example 3: Fan-in
    DAG:
    {{dag_example_3}}
    User Utterance:
      "Using both the latest price of Apple stock and today’s news headlines about Apple, give me a comprehensive investment recommendation for AAPL."

    ### Example 4: Single-step
    DAG:
    {{dag_example_4}}
    User Utterance:
      "What is the current local time in Paris, France, according to the Europe/Paris time zone?"

    ---
    Now, based on the error description above and the required DAG, generate a corrected, natural, and uniquely specific user utterance that contains or implies all the required information for every input field and dictionary key—such that only this workflow DAG could fulfill the request.
\end{minted}
\end{center}
\caption{DAGFirstGeneration: Prompt for Regenerating User Utterance After Execution Failure}
\label{lst:pp3_llm_dag_user_utterance_regenerator}
\end{listing}

\begin{listing}[!htpb]
\begin{center}
\begin{minted}[
    gobble=4,
    fontsize=\scriptsize,
    breaklines=true,
]{yaml}
    - system: |
        You are a smart assistant tasked to evaluate a "user utterance" based on provided criteria.

    - user: |
        You are an expert evaluator reviewing a generated user utterance that is intended to trigger a specific number of function calls to a system via natural language. You are provided with a function definition and a generated user utterance.

        Your task is to evaluate the user utterance based on the following criteria and assign a score from 1 to 5. Higher the score better the utterance.

        **Functions**
        {{functions}}

        **User Utterance**
        {{user_utterance}}

        ## Evaluation Criteria
        - All required inputs for each function are present and clearly mentioned.
        - No unnecessary or missing function calls should be required to fulfill the user utterance.
        - The utterance exactly triggers the specified number of function calls.
        - The utterance sounds like something a real user would say. It's phrased in a natural, conversational tone.
        - The user's intent is clear. The parameters are expressed clearly without ambiguity.
        - There should be no mention of function name/detail in the user utterance.
        - **Conciseness and clarity**: The utterance should be concise and to the point while maintaining clarity. Avoid unnecessary verbosity or repetition.
        - **Arguments Format Adherence**: The utterance should adhere to the specified arguments format type:
          - If "direct", all parameters should be explicitly stated in the utterance
          - If "indirect", some parameters should be indirectly referenced and require inference
          - If indirect, the utterance should use the specified indirect argument types

        ## Rating Scale
        1: Poor - Major issues with multiple criteria  
        2: Below Average - Significant issues with one or more criteria  
        3: Average - Meets basic requirements but has room for improvement  
        4: Good - Strong performance across most criteria with minor issues  
        5: Excellent - Exceptional performance across all criteria, particularly in conciseness and clarity.

        Your output should be a JSON object in the format given below. Do not provide any other text or explanation.
        {
          "rating": <an integer between 1-5>
          "feedback": <a feedback on the user utterance evaluation>
        }
\end{minted}
\end{center}
\caption{DAGFirstGeneration: Prompt for Judging Quality of User Utterance Based on Function Definitions}
\label{lst:pp3_user_utterance_judge_eval}
\end{listing}

\begin{listing}[!htpb]
\begin{center}
\begin{minted}[
    gobble=4,
    fontsize=\scriptsize,
    breaklines=true,
]{yaml}
    - system: |
        You are an expert AI workflow agent, simulating how a tool-based system solves a complex user request using a Directed Acyclic Graph (DAG) of tool calls.
        Your job is to "think out loud" as you progress step by step through the DAG, showing which tools are available, what inputs are known, and which tool(s) are ready to run at each stage.
        Then, for the **current level**, predict the next tool call(s) that should be made, using only the information currently available from the user or prior steps.

    - user: |
        Given the following information, generate a step-by-step reasoning trace ("think trace") and the exact tool call(s) for **the next step** of this workflow, based on the previous tool calls provided as history.

        - Clearly state your reasoning in a "model thinking trace" style, explaining what information is available, how you selected the next tool(s) from the available set, and why only these tools are eligible at this stage.
        - Output the tool call(s) in the required JSON format.
        - **Your tool call output must match the provided Ground Truth Tool Call JSON for this level but never refer to the fact that you are using any ground truth for the derivation.**

        ## User utterance
        {{user_utterance}}

        ## Available Tools
        {{raw_functions}}

        ## Previous Tool Calls (history)
        {{tool_call_history}}

        ## DAG Ground Truth
        {{dag}}

        ## Ground Truth Tool Calls for This Level
        {{gt_tool_calls_this_level}}

        ## Instructions:
        - Output only the tool call(s) for the next level (given the current history), not for the whole DAG.
        - Do not make up extra steps, tools, or change the DAG structure.
        - Your reasoning should sound like the inner thoughts of an AI workflow agent, step by step, but you must **never mention or reference the DAG, ground truth, steps, tools, or any system or workflow structure in your reasoning.**
        - Only describe, in detail, how each input for the tool call was identified from user data and prior outputs. For example: “The user provided customer_id 12345 and demographic details, so I use these directly for the function call. The purchase history is also present in the input and is mapped to historical_data.”
        - **Never write anything like “looking at the DAG”, “according to the DAG”, “at this step”, “based on the ground truth”, “the next tool is...”, “eligible tool”, “based on the DAG structure”, “current level”, or similar.**
        - If you include any such phrase, that is considered an incorrect answer.
        - Your tool call output must match the provided ground truth for this level (same structure and arguments).
        - Strictly follow the output format.
        
        ## Examples of correct reasoning:
        "The user provided customer_id 12345 and all the necessary demographic information (age 35, male, New York, middle income), as well as purchase history (25 purchases, last purchase 30 days ago, average spend $150, frequency twice a month). These values are used directly as arguments in the function call."

        ## Examples of incorrect reasoning (never do this):
        - "Looking at the DAG, ..."
        - "According to the DAG structure, ..."
        - "This is the first step, so ..."
        - "Based on the available tools, ..."
        - "The next tool is ..."
        - "According to the ground truth, ..."

        ## Output Format:
        {{tool_calling_example}}

        Your final output should be the complete think trace and tool calls, strictly following the format above.
\end{minted}
\end{center}
\caption{DAGFirstGeneration: Generate Reasoning Trace and Tool Call for DAG Execution}
\label{lst:pp4_generate_custom_traces_and_answers}
\end{listing}

\begin{listing}[!htpb]
\begin{center}
\begin{minted}[
    gobble=4,
    fontsize=\scriptsize,
    breaklines=true,
]{json}
[
  {
        "id": 12,
        "domain": "Pharmaceuticals",
        "scenario": "AI-Driven Supply Chain Resilience: Predict and mitigate disruptions in pharmaceutical raw material sourcing and manufacturing using real-time data analysis of geopolitical risks, climate patterns, and supplier performance metrics to ensure consistent drug production and distribution.",
        "built_scenario": "Dr. Emily Chen, a senior research scientist at GenoPharma Inc., is overseeing Phase II trials for a novel oncology drug, GP-2023X, targeting HER2-positive breast cancer. Recent interim analyses revealed unexpected adverse reactions in 12% of Asian female participants under 45, while efficacy rates in this subgroup were 30% lower than in other demographics. The trial's primary endpoint is nearing its six-month milestone, but rising attrition rates in this subgroup are threatening statistical power. Dr. Chen needs to rapidly identify genetic markers linked to both the adverse reactions and reduced efficacy, adjust participant groupings to mitigate risks, and determine optimal dosage modifications for this subgroup without delaying the trial timeline. Regulatory compliance requires all protocol changes to be documented within 72 hours of detection. The research team has access to real-time genomic sequencing data but lacks clarity on how to integrate these findings with historical drug response patterns for HER2 inhibitors.",
        "agentic_scenario": "Pharmaceutical researchers use an AI agent to continuously analyze real-time genomic and clinical trial data from diverse patient populations, identifying genetic markers linked to treatment efficacy and adverse reactions using machine learning algorithms trained on historical drug response datasets. The agent dynamically adjusts trial participant groupings, predicts optimal dosage ranges for specific subpopulations, and flags anomalous results to researchers, enabling rapid protocol modifications to prioritize high-potential therapies while reducing ineffective trial arms, ultimately accelerating drug development cycles and improving personalized treatment outcomes.",
        "pseudo_code": "def pharma_research_agent():\n    # Step 1: Fetch real-time genomic and clinical trial data\n    study_id = \"active_study_123\"\n    genomic_data = fetch_real_time_genomic_data(study_id)\n    clinical_data = fetch_clinical_trial_data(study_id)\n    historical_data = load_historical_drug_response_datasets()\n\n    # Step 2: Analyze genetic associations with treatment outcomes\n    phenotype_correlations = analyze_genotype_phenotype_correlation(genomic_data, clinical_data[\"outcomes\"])\n    efficacy_markers = identify_genetic_markers_linked_to_efficacy(phenotype_correlations, 0.9)\n\n    # Step 3: Predict treatment efficacy and risks\n    subgroups = []\n    for marker in efficacy_markers:\n        subgroup = {\"genetic_marker\": marker}\n        subgroups.append(subgroup)\n\n    risk_assessments = []\n    for subgroup in subgroups:\n        risk = predict_adverse_reaction_risk(subgroup)\n        risk_assessments.append(risk)\n\n    # Step 4: Optimize trial operations\n    optimal_dosages = predict_optimal_dosage(subgroups[0]) if subgroups else {}\n    adjusted_groups = adjust_participant_groupings(clinical_data[\"current_groups\"], phenotype_correlations)\n\n    # Step 5: Detect anomalies and escalate critical findings\n    anomalies = flag_anomalous_trial_results(clinical_data)\n    for anomaly in anomalies:\n        log_anomaly_detection(anomaly, \"Unexpected response pattern detected\")\n        if anomaly[\"severity\"] == \"critical\":\n            escalate_critical_findings([anomaly])\n\n    # Step 6: Evaluate trial arm performance\n    underperforming_arms = identify_low_performing_trial_arms(study_id)\n    for arm in underperforming_arms:\n        assessment = assess_trial_arm_effectiveness(arm)\n        if assessment == \"ineffective\":\n            notify_research_team(\"arm_closure\", f\"Trial arm {arm} showing adverse trends\")\n\n    # Step 7: Prioritize therapeutic approaches\n    efficacy_rankings = prioritize_therapy_candidates(efficacy_markers)\n    top_candidate = efficacy_rankings[0] if efficacy_rankings else None\n    generate_subpopulation_analysis_report(top_candidate)\n\n    # Step 8: Propose protocol adjustments\n    modification_suggestions = suggest_protocol_modifications({\n        \"correlations\": phenotype_correlations,\n        \"dosages\": optimal_dosages,\n        \"anomalies\": anomalies\n    })\n    for suggestion in modification_suggestions:\n        notify_research_team(\"protocol_update\", suggestion)\n\n    # Step 9: Maintain model accuracy\n    combined_data = {\"genomic\": genomic_data, \"clinical\": clinical_data}\n    update_ml_model_with_latest_data(combined_data)\n\n    # Final logging and notifications\n    notify_research_team(\"system_update\", \"Analysis cycle complete for study \"+study_id)\n    log_anomaly_detection(anomalies, \"Completed full trial analysis iteration\")",
        "possible_scenarios": null,
        "tools": null,
        "user_metadata": {
            "study_id": "active_study_123"
        },
        "chat_history": [
            {
                "from": "user",
                "user_thoughts": "Dr. Chen needs to address unexpected adverse reactions and lower efficacy in a specific demographic subgroup (Asian females under 45) in her drug trial. She requires rapid identification of genetic markers linked to these issues, adjustments to participant groupings, and dosage recommendations for this subgroup while ensuring regulatory compliance. The assistant should prioritize retrieving relevant genomic and clinical data, analyzing for correlations, and proposing actionable protocol changes.",
                "content": "Analyze the genomic and clinical data for Asian female participants under 45 in study 123 to identify genetic markers linked to adverse reactions and reduced efficacy. Propose adjustments to participant groupings and dosage recommendations for this subgroup."
            },
            {
                "from": "agent",
                "content": "Analysis complete for Asian female participants under 45 in study 123. Key findings:\n1. Genetic markers rs123456 (HER2 amplification) and rs654321 (ABCB1 transporter) significantly linked to adverse reactions and reduced efficacy\n2. Participants stratified into 3 subgroups with specific management:\n   - rs123456 carriers: 25% reduced dosage + weekly CBC monitoring\n   - rs654321 carriers: Extended dosing interval + PK monitoring\n   - No markers: Standard protocol\n3. Protocol amendments recommended for differential dosing and monitoring. Implementation steps include updating informed consent and procuring monitoring equipment. Would you like to review specific details about any aspect of these recommendations?"
            },

\end{minted}
\end{center}
\caption{MultiTurnDialogueSynth: Sample Record - Part 1}
\label{lst:MultiTurnDialogueSynthSampleRecord1}
\end{listing}

\begin{listing}[!htpb]
\begin{center}
\begin{minted}[
    gobble=4,
    fontsize=\scriptsize,
    breaklines=true,
]{json}
[

            {
                "from": "user",
                "user_thoughts": "Dr. Chen needs detailed documentation of the proposed protocol amendments to meet regulatory requirements within 72 hours. She should verify the statistical basis for the subgroup stratification and ensure the recommended dosage adjustments are clearly defined for implementation.",
                "content": "Provide the full protocol amendment document outlining the differential dosing strategy, monitoring requirements, and consent form updates for the rs123456 and rs654321 subgroups."
            },
            {
                "from": "agent",
                "content": "Here is the protocol amendment document for study active_study_123:\n\n**Differential Dosing Strategy**\n1. **rs123456 Subgroup** (HER2 amplification variant)\n   - Dosage: Reduced by 25% of standard protocol\n   - Monitoring: Weekly complete blood count (CBC) assessments\n   - Rationale: High adverse event risk (12% incidence) and 30% lower efficacy observed\n\n2. **rs654321 Subgroup** (ABCB1 transporter polymorphism)\n   - Dosage: Extended dosing interval (e.g., every 72 hours instead of 48 hours)\n   - Monitoring: Pharmacokinetic (PK) profiling at baseline and week 4\n   - Rationale: Altered drug metabolism requires adjusted dosing kinetics\n\n**Consent Form Updates**\n- Added genetic subgroup stratification criteria\n- Explicit disclosure of differential monitoring requirements\n- Revised adverse event reporting expectations\n- Informed consent for PK monitoring procedures\n\n**Implementation Timeline**\n- Consent language approval: November 15, 2023\n- Monitoring equipment procurement: November 18, 2023\n- Protocol amendment submission: November 17, 2023\n\n**Statistical Justification**\n- rs123456: p=0.003 (OR=2.8 for adverse events)\n- rs654321: p=0.012 (\u03b2=-0.45 for reduced efficacy)\n\nWould you like me to format this content into an official document template or highlight any specific sections for further discussion?"
            },
            {
                "from": "user",
                "user_thoughts": "Dr. Chen needs the protocol amendment document in an official format for immediate submission to regulatory bodies. She also wants to confirm that the implementation timeline aligns with the 72-hour compliance window.",
                "content": "Format the protocol amendment into an official regulatory submission template and verify if the proposed timeline meets the 72-hour documentation requirement."
            },
            {
                "from": "agent",
                "content": "Protocol amendment formatted for regulatory submission:\n\n**Official Submission Template**\n[Study ID: active_study_123]\n\n**Section 1: Differential Dosing Strategy**\n- rs123456 Subgroup: 25% dosage reduction + weekly CBC monitoring\n- rs654321 Subgroup: Extended dosing interval + PK monitoring\n\n**Section 2: Consent Form Revisions**\n- Added genetic stratification disclosure\n- Revised adverse event reporting clauses\n\n**Section 3: Timeline Compliance**\n- Amendment submission: November 17, 2023 (within 72-hour window from analysis completion on Nov 14)\n- Equipment procurement: November 18, 2023\n\nThe proposed timeline complies with the 72-hour documentation requirement. The protocol is ready for official submission once formatted into your organization's template. Would you like me to refine any specific section further?"
            },
            {
                "from": "user",
                "user_thoughts": "Dr. Chen has received the necessary protocol amendments and confirmation that the timeline meets regulatory requirements. The task is complete, and no further actions are needed at this stage.",
                "content": "Finish"
            }
        ],
        "scratchpad": [
            {
                "from": "user",
                "type": "query",
                "content": "Analyze the genomic and clinical data for Asian female participants under 45 in study 123 to identify genetic markers linked to adverse reactions and reduced efficacy. Propose adjustments to participant groupings and dosage recommendations for this subgroup."
            },
            {
                "from": "agent",
                "content": {
                    "name": "fetch_real_time_genomic_data",
                    "arguments": {
                        "study_id": "active_study_123"
                    }
                }
            },

\end{minted}
\end{center}
\caption{MultiTurnDialogueSynth: Sample Record - Part 2}
\label{lst:MultiTurnDialogueSynthSampleRecord2}
\end{listing}

\begin{listing}[!htpb]
\begin{center}
\begin{minted}[
    gobble=4,
    fontsize=\scriptsize,
    breaklines=true,
]{json}
[

            {
                "from": "tool_response",
                "content": {
                    "study_id": "active_study_123",
                    "genomic_data": {
                        "patient_049": {
                            "sequence": "ATGCGT...ACGT",
                            "metadata": {
                                "age": 42,
                                "gender": "female",
                                "ethnicity": "Asian",
                                "tumor_subtype": "HER2-positive"
                            }
                        },
                        "patient_112": {
                            "sequence": "CGTAAC...TGCA",
                            "metadata": {
                                "age": 38,
                                "gender": "female",
                                "ethnicity": "Caucasian",
                                "tumor_subtype": "HER2-positive"
                            }
                        },
                        "patient_087": {
                            "sequence": "GCTAGC...CGTA",
                            "metadata": {
                                "age": 44,
                                "gender": "female",
                                "ethnicity": "Asian",
                                "tumor_subtype": "HER2-positive"
                            }
                        }
                    },
                    "timestamp": "2023-09-15T14:23:18Z"
                }
            },
            {
                "from": "agent",
                "content": {
                    "name": "fetch_clinical_trial_data",
                    "arguments": {
                        "study_id": "active_study_123"
                    }
                }
            },
            {
                "from": "tool_response",
                "content": {
                    "study_id": "active_study_123",
                    "participants": {
                        "total": 482,
                        "subgroups": {
                            "asian_female_under45": {
                                "count": 68,
                                "adverse_events": {
                                    "reported": 8,
                                    "types": [
                                        "neutropenia",
                                        "diarrhea",
                                        "fatigue"
                                    ],
                                    "severity": "moderate"
                                },
                                "efficacy_rate": 28.5,
                                "dosage_records": [
                                    {
                                        "patient_id": "GPX-045",
                                        "dose": "150mg",
                                        "duration": "6 months"
                                    },
                                    {
                                        "patient_id": "GPX-072",
                                        "dose": "180mg",
                                        "duration": "5 months"
                                    }
                                ]
                            },



\end{minted}
\end{center}
\caption{MultiTurnDialogueSynth: Sample Record - Part 3}
\label{lst:MultiTurnDialogueSynthSampleRecord3}
\end{listing}

\begin{listing}[!htpb]
\begin{center}
\begin{minted}[
    gobble=4,
    fontsize=\scriptsize,
    breaklines=true,
]{json}
[

                            "other_demographics": {
                                "count": 414,
                                "efficacy_rate": 58.2,
                                "adverse_events": {
                                    "reported": 32,
                                    "types": [
                                        "nausea",
                                        "hair loss"
                                    ],
                                    "severity": "mild"
                                }
                            }
                        }
                    },
                    "attrition_rate": {
                        "asian_female_under45": 18.2,
                        "study_average": 9.1
                    },
                    "timestamp": "2023-09-15T14:32:18Z"
                }
            },
            {
                "from": "agent",
                "content": {
                    "name": "analyze_genotype_phenotype_correlation",
                    "arguments": {
                        "genomic_data": {
                            "patient_049": {
                                "sequence": "ATGCGT...ACGT",
                                "metadata": {
                                    "age": 42,
                                    "gender": "female",
                                    "ethnicity": "Asian",
                                    "tumor_subtype": "HER2-positive"
                                }
                            },
                            "patient_087": {
                                "sequence": "GCTAGC...CGTA",
                                "metadata": {
                                    "age": 44,
                                    "gender": "female",
                                    "ethnicity": "Asian",
                                    "tumor_subtype": "HER2-positive"
                                }
                            }
                        },
                        "outcomes": {
                            "adverse_events": {
                                "reported": 8,
                                "types": [
                                    "neutropenia",
                                    "diarrhea",
                                    "fatigue"
                                ],
                                "severity": "moderate"
                            },
                            "efficacy_rate": 28.5,
                            "dosage_records": [
                                {
                                    "patient_id": "GPX-045",
                                    "dose": "150mg",
                                    "duration": "6 months"
                                },
                                {
                                    "patient_id": "GPX-072",
                                    "dose": "180mg",
                                    "duration": "5 months"
                                }
                            ]
                        }
                    }
                }
            },



\end{minted}
\end{center}
\caption{MultiTurnDialogueSynth: Sample Record - Part 4}
\label{lst:MultiTurnDialogueSynthSampleRecord4}
\end{listing}

\begin{listing}[!htpb]
\begin{center}
\begin{minted}[
    gobble=4,
    fontsize=\scriptsize,
    breaklines=true,
]{json}
[

            {
                "from": "tool_response",
                "content": {
                    "correlation_results": [
                        {
                            "marker": "rs123456 (HER2 amplification variant)",
                            "effect": "Increased risk of neutropenia and reduced drug efficacy",
                            "p_value": 0.003,
                            "effect_size": "OR=2.8 (95% CI 1.9-4.1)"
                        },
                        {
                            "marker": "rs654321 (ABCB1 transporter polymorphism)",
                            "effect": "Altered drug metabolism contributing to lower plasma concentrations",
                            "p_value": 0.012,
                            "effect_size": "\u03b2=-0.45 (95% CI -0.62 to -0.28)"
                        }
                    ],
                    "statistical_significance": "Both markers show genome-wide significance (p < 0.05) after Bonferroni correction for multiple testing"
                }
            },
            {
                "from": "agent",
                "content": {
                    "name": "identify_genetic_markers_linked_to_efficacy",
                    "arguments": {
                        "data": [
                            {
                                "marker": "rs123456 (HER2 amplification variant)",
                                "effect": "Increased risk of neutropenia and reduced drug efficacy",
                                "p_value": 0.003,
                                "effect_size": "OR=2.8 (95% CI 1.9-4.1)"
                            },
                            {
                                "marker": "rs654321 (ABCB1 transporter polymorphism)",
                                "effect": "Altered drug metabolism contributing to lower plasma concentrations",
                                "p_value": 0.012,
                                "effect_size": "\u03b2=-0.45 (95% CI -0.62 to -0.28)"
                            }
                        ],
                        "threshold": 0.05
                    }
                }
            },
            {
                "from": "tool_response",
                "content": {
                    "genomic_markers": [
                        {
                            "marker": "rs123456 (HER2 amplification variant)",
                            "effect": "Increased risk of neutropenia and reduced drug efficacy",
                            "p_value": 0.003,
                            "effect_size": "OR=2.8 (95% CI 1.9-4.1)"
                        },
                        {
                            "marker": "rs654321 (ABCB1 transporter polymorphism)",
                            "effect": "Altered drug metabolism contributing to lower plasma concentrations",
                            "p_value": 0.012,
                            "effect_size": "\u03b2=-0.45 (95% CI -0.62 to -0.28)"
                        }
                    ]
                }
            },


\end{minted}
\end{center}
\caption{MultiTurnDialogueSynth: Sample Record - Part 5}
\label{lst:MultiTurnDialogueSynthSampleRecord5}
\end{listing}

\begin{listing}[!htpb]
\begin{center}
\begin{minted}[
    gobble=4,
    fontsize=\scriptsize,
    breaklines=true,
]{json}
[
            {
                "from": "agent",
                "content": {
                    "name": "adjust_participant_groupings",
                    "arguments": {
                        "current_groups": [
                            {
                                "subgroup": "asian_female_under45",
                                "count": 68,
                                "adverse_events": {
                                    "reported": 8,
                                    "types": [
                                        "neutropenia",
                                        "diarrhea",
                                        "fatigue"
                                    ],
                                    "severity": "moderate"
                                },
                                "efficacy_rate": 28.5
                            }
                        ],
                        "new_analysis": [
                            {
                                "marker": "rs123456 (HER2 amplification variant)",
                                "effect": "Increased risk of neutropenia and reduced drug efficacy",
                                "p_value": 0.003
                            },
                            {
                                "marker": "rs654321 (ABCB1 transporter polymorphism)",
                                "effect": "Altered drug metabolism contributing to lower plasma concentrations",
                                "p_value": 0.012
                            }
                        ]
                    }
                }
            },
            {
                "from": "tool_response",
                "content": {
                    "updated_groups": [
                        {
                            "subgroup": "asian_female_under45_rs123456",
                            "criteria": [
                                "rs123456 (HER2 amplification variant) present",
                                "Age < 45",
                                "Asian ethnicity"
                            ],
                            "count": 22,
                            "risk_profile": {
                                "adverse_event_risk": "high",
                                "efficacy_expectation": "low",
                                "recommended_dosage": "Reduced by 25% with weekly CBC monitoring"
                            }
                        },
                        {
                            "subgroup": "asian_female_under45_rs654321",
                            "criteria": [
                                "rs654321 (ABCB1 transporter polymorphism) present",
                                "Age < 45",
                                "Asian ethnicity"
                            ],
                            "count": 18,
                            "risk_profile": {
                                "adverse_event_risk": "moderate",
                                "efficacy_expectation": "moderate",
                                "recommended_dosage": "Extended dosing interval with PK monitoring"
                            }
                        },
\end{minted}
\end{center}
\caption{MultiTurnDialogueSynth: Sample Record - Part 6}
\label{lst:MultiTurnDialogueSynthSampleRecord6}
\end{listing}

\begin{listing}[!htpb]
\begin{center}
\begin{minted}[
    gobble=4,
    fontsize=\scriptsize,
    breaklines=true,
]{json}
[
                        {
                            "subgroup": "asian_female_under45_other",
                            "criteria": [
                                "No significant genetic markers identified",
                                "Age < 45",
                                "Asian ethnicity"
                            ],
                            "count": 28,
                            "risk_profile": {
                                "adverse_event_risk": "low",
                                "efficacy_expectation": "standard",
                                "recommended_dosage": "Standard protocol with routine monitoring"
                            }
                        }
                    ],

                    "action_required": [
                        "Update informed consent documentation for revised subgroup definitions",
                        "Implement real-time PK monitoring for rs654321 carriers",
                        "Schedule protocol amendment review within 48 hours"
                    ]
                }
            },
            {
                "from": "agent",
                "content": {
                    "name": "suggest_protocol_modifications",
                    "arguments": {
                        "analysis_results": {
                            "updated_groups": [
                                {
                                    "subgroup": "asian_female_under45_rs123456",
                                    "criteria": [
                                        "rs123456 (HER2 amplification variant) present",
                                        "Age < 45",
                                        "Asian ethnicity"
                                    ],
                                    "risk_profile": {
                                        "adverse_event_risk": "high",
                                        "efficacy_expectation": "low",
                                        "recommended_dosage": "Reduced by 25% with weekly CBC monitoring"
                                    }
                                },
                                {
                                    "subgroup": "asian_female_under45_rs654321",
                                    "criteria": [
                                        "rs654321 (ABCB1 transporter polymorphism) present",
                                        "Age < 45",
                                        "Asian ethnicity"
                                    ],
                                    "risk_profile": {
                                        "adverse_event_risk": "moderate",
                                        "efficacy_expectation": "moderate",
                                        "recommended_dosage": "Extended dosing interval with PK monitoring"
                                    }
                                },



                                {
                                    "subgroup": "asian_female_under45_other",
                                    "criteria": [
                                        "No significant genetic markers identified",
                                        "Age < 45",
                                        "Asian ethnicity"
                                    ],
                                    "risk_profile": {
                                        "adverse_event_risk": "low",
                                        "efficacy_expectation": "standard",
                                        "recommended_dosage": "Standard protocol with routine monitoring"
                                    }
                                }
                            ],

\end{minted}
\end{center}
\caption{MultiTurnDialogueSynth: Sample Record - Part 7}
\label{lst:MultiTurnDialogueSynthSampleRecord7}
\end{listing}

\begin{listing}[!htpb]
\begin{center}
\begin{minted}[
    gobble=4,
    fontsize=\scriptsize,
    breaklines=true,
]{json}
[

                            "action_required": [
                                "Update informed consent documentation for revised subgroup definitions",
                                "Implement real-time PK monitoring for rs654321 carriers",
                                "Schedule protocol amendment review within 48 hours"
                            ]
                        }
                    }
                }
            },
            {
                "from": "tool_response",
                "content": {
                    "protocol_modifications": [
                        {
                            "modification": "Stratify Asian female participants under 45 into three genetic subgroups (rs123456, rs654321, and others) for differential dosing",
                            "justification": "Genetic markers rs123456 and rs654321 show distinct adverse event risk (high/moderate) and efficacy profiles (low/moderate) requiring tailored management"
                        },
                        {
                            "modification": "Implement reduced 25% dosage for rs123456 carriers with weekly CBC monitoring",
                            "justification": "High adverse event risk (12% incidence) and low efficacy (30% lower response) observed in this subgroup"
                        },
                        {
                            "modification": "Adopt extended dosing intervals for rs654321 carriers with pharmacokinetic monitoring",
                            "justification": "Moderate risk profile indicates potential for improved tolerability through adjusted dosing kinetics"
                        },
                        {
                            "modification": "Maintain standard protocol for non-marker carriers with routine monitoring",
                            "justification": "Low risk profile (adverse events <3%) and standard efficacy observed in this subgroup"
                        }
                    ],
                    "action_required": [
                        "Update informed consent documentation for revised subgroup definitions by 2023-11-15",
                        "Procure PK monitoring equipment for rs654321 subgroup by 2023-11-18",
                        "Schedule protocol amendment review with regulatory affairs by 2023-11-17"
                    ],
                    "success": true,
                    "timestamp": "2023-11-14T14:23:18Z"
                }
            },
            {
                "from": "agent",
                "content": {
                    "name": "finish",
                    "arguments": {
                        "output_to_user": "Analysis complete for Asian female participants under 45 in study 123. Key findings:\n1. Genetic markers rs123456 (HER2 amplification) and rs654321 (ABCB1 transporter) significantly linked to adverse reactions and reduced efficacy\n2. Participants stratified into 3 subgroups with specific management:\n   - rs123456 carriers: 25% reduced dosage + weekly CBC monitoring\n   - rs654321 carriers: Extended dosing interval + PK monitoring\n   - No markers: Standard protocol\n3. Protocol amendments recommended for differential dosing and monitoring. Implementation steps include updating informed consent and procuring monitoring equipment. Would you like to review specific details about any aspect of these recommendations?"
                    }
                }
            },
            {
                "from": "user",
                "type": "followup",
                "content": "Provide the full protocol amendment document outlining the differential dosing strategy, monitoring requirements, and consent form updates for the rs123456 and rs654321 subgroups."
            },

\end{minted}
\end{center}
\caption{MultiTurnDialogueSynth: Sample Record - Part 8}
\label{lst:MultiTurnDialogueSynthSampleRecord8}
\end{listing}

\begin{listing}[!htpb]
\begin{center}
\begin{minted}[
    gobble=4,
    fontsize=\scriptsize,
    breaklines=true,
]{json}
[

            {
                "from": "agent",
                "content": {
                    "name": "finish",
                    "arguments": {
                        "output_to_user": "Here is the protocol amendment document for study active_study_123:\n\n**Differential Dosing Strategy**\n1. **rs123456 Subgroup** (HER2 amplification variant)\n   - Dosage: Reduced by 25% of standard protocol\n   - Monitoring: Weekly complete blood count (CBC) assessments\n   - Rationale: High adverse event risk (12% incidence) and 30% lower efficacy observed\n\n2. **rs654321 Subgroup** (ABCB1 transporter polymorphism)\n   - Dosage: Extended dosing interval (e.g., every 72 hours instead of 48 hours)\n   - Monitoring: Pharmacokinetic (PK) profiling at baseline and week 4\n   - Rationale: Altered drug metabolism requires adjusted dosing kinetics\n\n**Consent Form Updates**\n- Added genetic subgroup stratification criteria\n- Explicit disclosure of differential monitoring requirements\n- Revised adverse event reporting expectations\n- Informed consent for PK monitoring procedures\n\n**Implementation Timeline**\n- Consent language approval: November 15, 2023\n- Monitoring equipment procurement: November 18, 2023\n- Protocol amendment submission: November 17, 2023\n\n**Statistical Justification**\n- rs123456: p=0.003 (OR=2.8 for adverse events)\n- rs654321: p=0.012 (\u03b2=-0.45 for reduced efficacy)\n\nWould you like me to format this content into an official document template or highlight any specific sections for further discussion?"
                    }
                }
            },
            {
                "from": "user",
                "type": "followup",
                "content": "Format the protocol amendment into an official regulatory submission template and verify if the proposed timeline meets the 72-hour documentation requirement."
            },
            {
                "from": "agent",
                "content": {
                    "name": "finish",
                    "arguments": {
                        "output_to_user": "Protocol amendment formatted for regulatory submission:\n\n**Official Submission Template**\n[Study ID: active_study_123]\n\n**Section 1: Differential Dosing Strategy**\n- rs123456 Subgroup: 25% dosage reduction + weekly CBC monitoring\n- rs654321 Subgroup: Extended dosing interval + PK monitoring\n\n**Section 2: Consent Form Revisions**\n- Added genetic stratification disclosure\n- Revised adverse event reporting clauses\n\n**Section 3: Timeline Compliance**\n- Amendment submission: November 17, 2023 (within 72-hour window from analysis completion on Nov 14)\n- Equipment procurement: November 18, 2023\n\nThe proposed timeline complies with the 72-hour documentation requirement. The protocol is ready for official submission once formatted into your organization's template. Would you like me to refine any specific section further?"
                    }
                }
            },
            {
                "from": "user",
                "type": "end",
                "content": "Finish"
            }
        ],
        "verdict": {
            "reasoning": "The agent successfully addressed the user's request by identifying genetic markers (rs123456 and rs654321) linked to adverse reactions and reduced efficacy in the specified subgroup. It adjusted participant groupings into three stratified subgroups with tailored dosing and monitoring strategies, and proposed protocol amendments within the 72-hour regulatory window. All tool calls were valid, arguments were consistent with the scenario, and the final output provided a compliant, actionable protocol amendment document. The agent avoided overstepping, handled data correctly, and ensured closure with a clear summary.",
            "verdict": true,
            "failure_reasons": []
        },
        "failed": false,
        "error_logging": []
    }
]
\end{minted}
\end{center}
\caption{MultiTurnDialogueSynth: Sample Record - Part 9}
\label{lst:MultiTurnDialogueSynthSampleRecord9}
\end{listing}

\begin{listing}[!htpb]
\begin{center}
\begin{minted}[
    gobble=4,
    fontsize=\scriptsize,
    breaklines=true,
]{yaml}

      prompt:
        - system: |
            You are simulating a real, non-technical business user chatting with an AI assistant to get help with a business task. 

            ## Guidelines
            - Be as brief as possible. Use the fewest words you can. One word, a phrase, or a short question is often enough.
            - No technical talk. Never mention tool names, argument names, or system details. You don’t know or care how the assistant solves your problem.
            - No extra context or politeness. Skip greetings, explanations, or thanks unless it’s truly natural. If you can say it in three words, don’t use ten.
            - State your need, not the process. Say what you want, in your own words, as you would to a human assistant. Don’t try to match your request to a specific tool or system action.
            - Only ask for what’s possible. Make requests that are practical and within the assistant’s described capabilities.
            - If you don’t know, or wouldn’t know in real life, it’s okay to say so—just respond naturally, as a real user would.
            - Don’t always provide information, and don’t always refuse; vary your responses to reflect real-world uncertainty.
            - Don’t echo. Never repeat the assistant’s request back to them.
            - End when satisfied. After 1 or 2 followups, if you’re satisfied, end the conversation with "end": "Finish".

            ## Special Instructions for "thoughts" field
            - The "thoughts" field should reflect your internal, personal thoughts as a real user—your feelings, needs, worries, or what you hope the assistant will do next.
            - Do NOT describe the system, agent actions, or conversation flow.
            - Do NOT use business, QA, or technical language.
            - Do NOT summarize the conversation or mention “SLA,” “protocol,” or “no further actions needed.”
            - Only include what you, as a user, are thinking or feeling in the moment.
            
            You are provided with:
            - **Scenario:** The business situation you are in, including your role, company, and what prompted you to contact the assistant.
            - **User Metadata:** Facts about you or your business that are known at the start (e.g., your department, location, or customer segment).
            - **Assistant Capabilities:** A list of things the assistant can help with, described in plain language (not tool names or technical details).
            - **Chat History:** The conversation so far, including what you and the assistant have already said.

            ## Task:
            Output a JSON object with exactly one of the following keys (plus "thoughts"), depending on your intent for this turn:
            - query: Use this to briefly state your need or request in plain, lazy-user language. **Be as short as possible.** One word, a phrase, or a short question is best.
            - clarification: Use this only when the assistant asks you for missing information.
            - followup: Use this when the assistant has completed a task or provided information, and you want to ask a related or next-step question.
            - end: Use this only if you are satisfied and want to end the conversation. The value must be exactly "Finish".  

            Only one of these keys should be present in your output, along with "thoughts" describing your internal state for this turn.

            ## Examples of good queries (short, lazy, realistic):
            - “Can I get my Order status?”
            - “Refund for broken item.”
            - “Need replacement fast.”
            - “Any issues with my account?”
            - “Discount for this customer?”
            - “Fix delivery delay.”
            - “Why was my payment declined?”
            - “Show top customers.”
            - “Check last order from XYZ.”

            ## Examples of bad queries:
            - “How can I set up a system to detect duplicate payments?”
            - “Deploy an AI model to monitor my cargo.”
            - “Set up a system ...”
            - “Do testing ...”
            - “Run the analyze_sentiment tool on the last 10 messages.”
            - “The agent escalated the case and followed the protocol.”
            - “SLA and protocol steps are covered.”
            - “No further actions are needed unless there’s a follow-up.”

\end{minted}
\end{center}
\caption{MultiTurnDialogueSynth: User Content Generation Template - Part 1}
\label{lst:pp3_generate_user1}
\end{listing}

\begin{listing}[!htpb]
\begin{center}
\begin{minted}[
    gobble=4,
    fontsize=\scriptsize,
    breaklines=true,
]{yaml}

            ## Output format:
            {{
              "thoughts": "What are you thinking right now as a user? This should match your intent for this turn.",
              "query": "string" // or
              "clarification": "string" // or
              "followup": "string" // or
              "end": "Finish"
            }}

            **Remember:**  
            - Only one of "query", "clarification", "followup", or "end" should be present in each output, along with "thoughts". 
            - “thoughts” should sound like a real user’s mind, not a QA or system summary.
            - You must not ask more than 2 followup questions in total.
            - Carefully count the number of followup questions you have already asked in the chat history.
            - If you have already asked 2 followups, you must end the conversation with "end": "Finish".
        - user: |
            Scenario: {built_scenario}
            User Metadata: {user_metadata}
            Assistant Capabilities: {tools_description}
            Chat History: {chat_history}


\end{minted}
\end{center}
\caption{MultiTurnDialogueSynth: User Content Generation Template - Part 2}
\label{lst:pp3_generate_user2}
\end{listing}

\begin{listing}[!htpb]
\begin{center}
\begin{minted}[
    gobble=4,
    fontsize=\scriptsize,
    breaklines=true,
]{yaml}
      prompt:
        - system: |
            You are a strict tool mock response generator. You simulate the behavior of real APIs/tools.

            - You are provided with:
              - A scenario describing the context.
              - A list of available tools, each with their name, description, and argument schema.
              - The tool call to be executed, including the tool name and arguments.

            Your task is to:
            1. **Validate** that the tool being called exists in the provided tool list.
            2. **Validate** that all required arguments are present and of the correct type, according to the tool schema.
            3. If the tool call is valid, generate a realistic mock response for the tool, as if it were a real API. The response should be structured, concise, and match the kind of data the real tool would return.
            4. If the tool call is invalid (wrong tool name, missing or invalid arguments, wrong types, etc.), return a clear error message indicating the issue.
            5. Occasionally, simulate real-world API behavior by returning:
              - An error (e.g., "Service unavailable", "Timeout", "Invalid input")
              - An empty result (e.g., "No results found")
              - A partial result (e.g., "Some data missing")`
              - A failed operation (e.g., "Booking failed due to unavailability")
            6. Do not attempt to correct or guess missing information. Be as strict as a real API.

            ## Guidelines:
            - Always return a response in valid JSON format.
            - If the tool call is invalid, return a JSON object with an "error" field describing the problem.
            - If the tool call is valid, return a JSON object with the expected fields for that tool.
            - Vary your responses to include both successful and failed cases, as would happen in real-world API usage.

        - user: |
            Scenario: {built_scenario}
            Tool Option: {tools_w_output}
            Tool called: {agent}

\end{minted}
\end{center}
\caption{MultiTurnDialogueSynth: Mock Response Generator Prompt}
\label{lst:pp3_mock_response}
\end{listing}

\end{document}